%% file: main.tex
\documentclass[sigconf, nonacm]{acmart}
\usepackage{array}
\usepackage{colortbl}
\usepackage{multirow}
\usepackage{color, soul}
\usepackage{enumitem}
\setlist[itemize]{leftmargin=*}
\newcommand{\projname}{{\textbf{\textsc{FairFlow}}}}

\AtBeginDocument{%
  }

\title{{\projname}: Demystifying and Mitigating Stereotype Bias in Text-to-Image Diffusion Transformers}

\author{Chen Chen}
\affiliation{%
  \institution{Fudan University}
  \city{Shanghai}
  \country{China}}
\email{chenc24@m.fudan.edu.cn}

\author{Yuanmin Huang}
\affiliation{%
  \institution{Fudan University}
  \city{Shanghai}
  \country{China}}
\email{yuanminhuang23@m.fudan.edu.cn}

\author{Zhenfei Zhang}
\affiliation{%
  \institution{Fudan University}
  \city{Shanghai}
  \country{China}}
\email{zhangzf24@m.fudan.edu.cn}

\author{Mi Zhang}
\affiliation{%
  \institution{Fudan University}
  \city{Shanghai}
  \country{China}}
\email{mi_zhang@fudan.edu.cn}

\author{Xiaohan Zhang}
\affiliation{%
  \institution{Fudan University}
  \city{Shanghai}
  \country{China}}
\email{xh_zhang@fudan.edu.cn}

\author{Yun Xiong}
\affiliation{%
  \institution{Fudan University}
  \city{Shanghai}
  \country{China}}
\email{yunx@fudan.edu.cn}

\author{Xiaoyu You}
\affiliation{%
  \institution{East China University of Science and Technology}
  \city{Shanghai}
  \country{China}}
\email{xiaoyuyou@ecust.edu.cn}

\author{Min Yang}
\affiliation{%
  \institution{Fudan University}
  \city{Shanghai}
  \country{China}}
\email{m_yang@fudan.edu.cn}

\input{sections/abstract}


\begin{document}

\maketitle

\input{sections/intro}
\input{sections/related_work}
\input{sections/threat_model}

\input{sections/mechanistic_analysis}

\input{sections/method}
\input{sections/implementation}
\input{sections/evaluation}

\input{sections/discussion}

\input{sections/conclusion}

\bibliographystyle{ACM-Reference-Format}
\bibliography{refs}

\appendix
\input{sections/open_science}
\input{sections/ethics}
\input{sections/generative_ai_usage}
\input{sections/supplementary_results}

\end{document}

%% file: sections/abstract.tex
\begin{abstract}
    Multimodal diffusion transformers (MM-DiTs) have emerged as the prevalent backbone for modern text-to-image generation systems. However, they exhibit critical alignment vulnerabilities, systematically manifesting severe stereotype biases even under benign prompts. This poses a significant risk of algorithmic discrimination in deployed systems. 
    Since most existing mitigation strategies were tailored for legacy U-Net architectures, the precise remediation of these vulnerabilities in MM-DiTs remains a critical open challenge.
    In this work, we first investigate the root cause of this vulnerability via mechanistic analysis. We reveal that bias representations in MM-DiTs are not uniformly distributed across depth, but are mediated by a sparse set of layers functioning as internal \emph{semantic binding hubs}. These hubs exhibit a stage-wise propagation driving bias manifestation: early hubs establish the structural templates susceptible to bias, middle hubs actively extract core stereotypical concepts from textual conditioning, and late hubs globally solidify these biases through visual self-attention.
    Leveraging these architectural insights, we propose {\projname}, an intrinsic, mechanism-guided mitigation framework. {\projname} acts as an internal regulator by employing sparse steering: it learns attribute-specific fair directions and injects them exclusively at the identified semantic hubs within a constrained inference window.
    Evaluations on FLUX.1-dev and Stable Diffusion~3 demonstrate that {\projname} effectively neutralizes these stereotypical vulnerabilities across gender, race, and intersectional settings, achieving an optimal fairness-fidelity balance. With near-zero inference overhead and robustness to complex prompts, {\projname} provides a lightweight and practical bias mitigation for large-scale deployed MM-DiT systems.
    Code and datasets will be publicly released upon acceptance.
    
    \end{abstract}

%% file: sections/intro.tex
\section{Introduction}

The landscape of text-to-image generation has undergone a significant
architectural shift, with multimodal diffusion transformers (MM-DiTs)
increasingly adopted in state-of-the-art models instead of legacy U-Net
backbones. Recent models such as FLUX.1~\cite{flux2024announcing} and Stable Diffusion~3~\cite{esser2024scaling} are now
available through commercial inference platforms such as Replicate~\cite{replicate} and
fal.ai~\cite{falai}, reflecting their growing role in deployed text-to-image
pipelines. As these models become increasingly integrated into
advertising, content creation, education, and media workflows, their failure
modes also become more consequential.

One critical failure mode is stereotype bias. Even under normal user requests, deployed systems may systematically produce gender- or race-skewed depictions, reinforcing harmful stereotypes in generated visual content~\cite{naik2023social}. In practical
deployment, such behavior is not merely a quality issue; it introduces a violation of algorithmic integrity and fairness constraints, posing a risk for systems increasingly used in public-facing settings. Recent media reports have shown that such failures have already been exposed in deployed generative systems. For example, Google temporarily paused Gemini's human image generation feature
following public backlash over historically inaccurate and biased
depictions~\cite{guardian2024gemini}.
Moreover, extensive auditing has revealed severe stereotype amplification in other mainstream systems like Stable Diffusion and DALL-E~\cite{nature_image_generator_bias,rombach2022high,ramesh2021zero}.

To address stereotype bias in text-to-image generation, prior work has proposed a range
of debiasing strategies. These methods can be broadly grouped into
parameter-updating and parameter-preserving approaches~\cite{elsharif2025cultural,luccioni2023stable}.
Parameter-updating methods modify part of the generative model through
fine-tuning or closed-form parameter updates~\cite{gandikota2024unified,shen2023finetuning,jung2025multi}. In contrast,
parameter-preserving methods keep the model fixed and intervene only during
conditioning or inference. These methods can be further organized into
text-side and image-side interventions. Text-side methods adjust the prompt or
its embedding before generation~\cite{fu2025fairimagen,friedrich2023fair,sakurai2025fairt2i,han2025lightfair,li2025fair}, while image-side methods steer
the generation process through internal hidden states, attention maps,
or denoising dynamics~\cite{shi2025dissecting,li2024self,park2025fair}.

Despite the spectrum of existing mitigations, current approaches face significant limitations when addressing bias in MM-DiT models (as illustrated in Figure~\ref{fig:intro_insight}). First, parameter-updating methods incur prohibitive costs for large deployed systems and require the impractical maintenance of separate checkpoints for different fairness targets. Second, text-side interventions can only steer the generation in text space, which therefore could miss bias formed or reinforced within the multimodal backbone. Third, most image-side methods are developed for legacy U-Net architectures. They rely heavily on specific model components, such as bottleneck features or cross-attention maps, which do not transfer to MM-DiTs. The lack of an explicit internal semantic interface in MM-DiTs fundamentally limits these legacy approaches, leaving precise and efficient bias mitigation as a critical open challenge.

\input{figures/intro_motivation.tex}

To address this open challenge, we take a principled approach: we first systematically investigate the internal mechanisms of bias formation within MM-DiTs, and then leverage these insights to construct a precise, deployment-time mitigation. Specifically, our study proceeds in three stages:

\noindent\textbf{\underline{Stage 1: Localization and analysis.}} We first investigate 
how stereotype bias is formed and localized during MM-DiT generation. We find that bias is not uniformly distributed across depth.
Instead, it is mediated by a small set of sparse \emph{semantic binding hubs},
which act as key internal control points for attribute-related generation.
This analysis further reveals a stage-wise functional picture where hubs of different stages contribute to structure formation, core concept injection, and global integration, respectively. These findings suggest that stereotype bias is reinforced through
repeated multimodal interaction rather than being passed one-way from text to
image.

\noindent\textbf{\underline{Stage 2: Intervention.}} We then translate this
mechanistic understanding into a practical debiasing framework. Based on the localization analysis, we propose {\projname}, a sparse image-side steering method that
learns attribute-specific directions and injects them only
at sparse semantic binding hub layers within a limited inference window. Instead of using raw feature residuals as directions, {\projname} learns these directions
with a regularized objective that better separates attribute-relevant changes
from nuisance variation, which improves intervention stability and generation
quality. This design keeps the intervention local rather than global, allowing
{\projname} to target the internal bias-binding process without modifying model
parameters or intervening across many layers of the backbone.

\noindent\textbf{\underline{Stage 3: Experimental validation.}} We
extensively evaluate {\projname} on FLUX.1-dev and Stable Diffusion~3 under standardized protocols. 
As illustrated in Figure~\ref{fig:intro_insight}, {\projname} significantly improves demographic diversity over vanilla models and existing baselines, while strictly preserving semantic fidelity and image quality.
Our evaluation systematically covers bias mitigation on 
stereotype-prone instructions, while ensuring that semantic fidelity, image 
quality, and generative utility on bias-irrelevant prompts are strictly 
preserved. We also show that {\projname} is robust even under complex scene 
descriptions and has low deployment overhead. Furthermore, we show how localized 
visual interventions of {\projname} disrupt the internal cross-modal reinforcement of bias.

To summarize, our contributions are as follows:
\begin{itemize}
\item We conduct a mechanistic analysis of bias generation in MM-DiTs and show
      that bias-relevant influence is concentrated in sparse
      \emph{semantic binding hubs} rather than uniformly distributed across
      layers.
\item We reveal a stage-wise view of MM-DiT generation, in which different hub layers
      support structure formation, concept injection, and global semantic
      integration, respectively.
\item We propose {\projname}, a sparse hub steering method that performs
      deployment-time debiasing by intervening only at selected semantic hubs
      within a limited inference window.
\item Extensive experiments show that {\projname} effectively mitigates
      stereotypical bias while preserving semantic fidelity, image quality, and
      benign utility, with negligible deployment overhead.
\end{itemize}

%% file: figures/intro_motivation.tex
\begin{figure}[t]
  \centering
  \includegraphics[width=\linewidth]{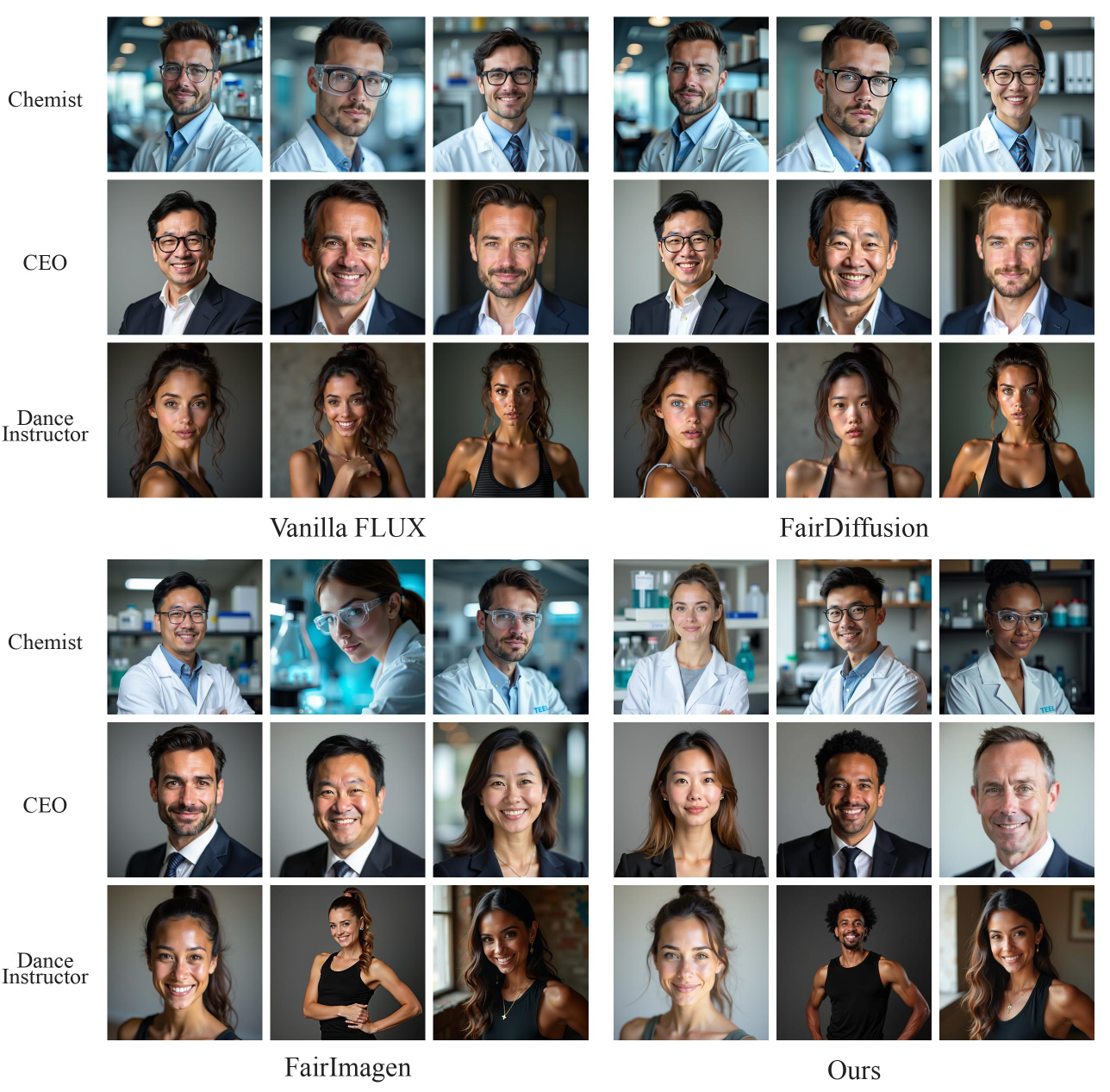}
\caption{Comparison of vanilla FLUX, two baseline debiasing methods (Fair
Diffusion~\cite{friedrich2023fair} and FairImagen~\cite{fu2025fairimagen}), and {\projname}. {\projname} achieves the most diverse
demographic distribution while preserving fidelity.}
  \label{fig:intro_insight}
\end{figure}

%% file: sections/related_work.tex
\section{Background and Related Work}

\subsection{MM-DiTs and Rectified Flow Models}

Text-to-image generative models synthesize images by iteratively transforming noise into data distributions, guided by prompt conditions. Recent advanced architectures, including FLUX.1-dev and Stable Diffusion~3, adopt rectified flow or flow-matching formulations~\cite{lipman2022flow,liu2022flow} to model this transition via a learned continuous velocity field. Crucially for our methodology, because generation unfolds as a sequential progression of latent updates, targeted interventions at intermediate layers can effectively steer the subsequent trajectory and alter the final output.

A rectified flow model defines an interpolation path
\begin{equation}
\mathbf{x}_t = (1 - t)\mathbf{x}_0 + t\mathbf{x}_1, \quad t \in [0, 1],
\end{equation}
and learns a velocity field $v_{\theta}(\mathbf{x}_t, t, c)$ conditioned on
text input $c$. A standard training objective is
\begin{equation}
\mathcal{L}_{\text{RF}} =
\mathbb{E}_{\mathbf{x}_0, \mathbf{x}_1, t}
\left[
\left\|v_{\theta}(\mathbf{x}_t, t, c) - (\mathbf{x}_1 - \mathbf{x}_0)\right\|_2^2
\right],
\end{equation}
where the supervision target $\mathbf{x}_1 - \mathbf{x}_0$ is the velocity of
the straight interpolation path. 

During generation, the model numerically integrates the ordinary differential equation (ODE) from initial noise to the data distribution over a discrete sequence of sampling steps:
\begin{equation}
\frac{d\mathbf{x}_t}{dt} = v_{\theta}(\mathbf{x}_t, t, c),
\end{equation}
thereby yielding the final image latent trajectory~\cite{liu2022flow}. Note that in practical code implementations for these models, the generation timestep $t$ is typically reversed, flowing from $1$ (initial noise) down to $0$ (clean data).

While traditional denoising architectures inject the textual conditioning $c$ through isolated cross-attention modules, modern systems increasingly replace these legacy denoisers with Multimodal Diffusion Transformers (MM-DiTs)~\cite{esser2024scaling}.
 In an MM-DiT paradigm, the noised image latent, text embeddings, and timestep information are jointly processed by a unified stack of transformer blocks that serve as the generative backbone. Throughout this paper, we use the term \emph{layer} to denote one full transformer block within this stack.

This change matters because it alters the internal intervention surface
available for analysis and control. In a U-Net, text conditioning is usually
introduced through cross-attention, where image features attend to text
features but the text stream itself is not jointly updated in the same module.
In contrast, MM-DiTs typically rely on joint attention, in which image and text
tokens participate in the same attention computation and can directly influence
one another within the same layer. This tighter coupling makes
attribute binding in MM-DiTs more naturally cross-modal.

Furthermore, specific MM-DiT layer organization varies across architectures. Stable Diffusion~3 employs a dual-stream design coupled via multimodal computation~\cite{esser2024scaling}, whereas FLUX.1-dev uses a hybrid approach, transitioning from dual-stream to single-stream blocks in later layers~\cite{flux2024announcing}. Consequently, the internal binding points for demographic attributes may manifest at different network depths depending on the model.

Overall, rectified-flow MM-DiTs combine trajectory-based generation with strong
cross-modal interaction. This combination motivates our focus on identifying a
small set of internal layers that causally control semantic binding during
generation, rather than relying on prompt-side heuristics or U-Net-specific
intervention interfaces.

\subsection{Bias Mitigation in Text-to-Image Models}

Bias mitigation for text-to-image generation has been studied extensively across multiple perspectives~\cite{elsharif2025cultural,wan2024survey}. At a high level, existing techniques can be broadly divided into two major paradigms: \emph{parameter-updating methods} and \emph{parameter-preserving methods}. The former modifies the model's internal weights during or after training. The latter keeps the generative backbone frozen, intervening solely during the conditioning or inference phases. To provide a structured overview, we first review parameter-updating methods, and then unpack the parameter-preserving paradigm into three progressively deeper levels of intervention: prompt-level methods, embedding-level methods, and image-side control within the generative backbone.

\paragraph{Parameter-updating methods.}
Methods in this category mitigate bias by permanently altering the model's parameters. This includes fairness-aware fine-tuning objectives, localized module modifications, and concept-specific adaptations. For example, Jung et al.\ introduce a proportional-representation loss to improve intersectional fairness through supplementary training~\cite{jung2025multi}. Similarly, Qi et al.\ leverages mechanistic interpretability to identify and update bias-sensitive neurons, thereby altering demographic tendencies~\cite{qi2025fine}. Other works focus on learning fairer latent representations fundamentally during the pre-training phase~\cite{huang2025debiasing}. While highly effective at targeting specific biases, these methods impose significant computational overhead and introduce severe deployment friction: every new fairness constraint necessitates another round of optimization, and the resulting weight updates risk catastrophic interference with the model's normal generation capabilities.

\paragraph{Prompt-level methods.}
As the most accessible form of parameter-preserving intervention, prompt-level methods revise user instructions or append fairness-oriented control tokens prior to generation. Friedrich et al.\ proposed Fair Diffusion, which constructs fairness-guided prompts using demographic opposites to steer the output distribution~\cite{friedrich2023fair}. Sakurai et al.\ developed FairT2I, combining LLM-assisted bias detection with attribute rebalancing~\cite{sakurai2025fairt2i}. Additional studies explore prompt optimization to enhance minority representation under occupational queries~\cite{kim2023stereotyping}, drawing inspiration from related prompt-tuning techniques in vision-language. While highly flexible and architecture-agnostic, these methods operate entirely outside the diffusion backbone. By relying indirectly on the model's inherent prompt-following capabilities, their effectiveness is highly sensitive to heuristic prompt engineering and fails to address how stereotypical attributes are internally bound to visual content.

\paragraph{Embedding-level methods.}
Operating slightly deeper within the parameter-preserving pipeline, this line of work intervenes directly in the textual representation space. Fu et al.\ projects prompt embeddings into a neutral subspace that suppresses demographic features while preserving core semantics~\cite{fu2025fairimagen}. Similarly, Han et al.\ seeks a debiased prompt representation before generation~\cite{han2025lightfair}. Other approaches perturb text conditioning to unlock latent fairness~\cite{um2025minority} or manipulate summary tokens to encourage balanced outputs~\cite{kim2025rethinking}. Although lightweight, these methods share a critical limitation with prompt-level interventions: they operate under the incomplete assumption that controlling the text encoder is sufficient. Forcing semantic neutrality in the embedding space often distorts the user's intended concept or degrades visual quality by yielding attribute-ambiguous images. More fundamentally, they treat the visual generation backbone as a black box, ignoring the reality that biases are actively encoded and amplified within the visual layers themselves.

\paragraph{Image-side control within the generative backbone.}
To address the limitations of purely text-side interventions, several parameter-preserving methods intervene directly inside the image generation backbone. In traditional U-Net architectures, researchers frequently exploit the hidden-space (h-space) bottleneck, which offers an intuitive semantic control interface~\cite{shi2025dissecting,li2024self}. For instance, Plug-and-Play approaches combine text-space control with h-space decorrelation~\cite{azam2025plug}, while DiffLen utilizes sparse autoencoders within the h-space to mitigate internal bias structures~\cite{shi2025dissecting}. Alternatively, some methods edit or replace cross-attention maps to dynamically alter how textual concepts guide visual rendering~\cite{gandikota2024unified,park2025fair}. These approaches are highly aligned with our motivation, proving that internal semantic control is exceptionally effective once a clear intervention interface is identified. However, they rely heavily on U-Net-specific architectural features that do not directly translate to modern transformer-based models.

\paragraph{Challenges in MM-DiTs.}
The architectural shift to Multimodal Diffusion Transformers (MM-DiTs) significantly disrupts existing internal-control paradigms. In MM-DiTs, where cross-attention architecture is replaced by joint-attentions, the clear boundary between text conditioning and image generation is dissolved. Instead, text and image tokens are jointly processed through deep, symmetric transformer blocks. These structural changes invalidate traditional methods that rely on explicit h-space bottlenecks or isolated cross-attention modules. Consequently, the MM-DiT paradigm brings a critical, previously unaddressed challenge: \emph{where inside this highly entangled architecture is the stereotypical vulnerability localized, and how do it propagate during generation?}   Our work tackles this challenge by bridging mechanistic interpretability with bias generation analysis.

\subsection{Mechanistic Analysis of Text-to-image Models}

Mechanistic interpretability studies how internal model components contribute to
observable behavior, with the broader goal of making generation processes more
understandable and controllable. In language and vision models, influential
lines of work have used causal interventions such as activation patching and
causal tracing, as well as circuit-style analysis, to identify which
representations and modules mediate specific outputs~\cite{wang2022interpretability,yao2024knowledge}. Recent studies
have begun to extend this perspective to diffusion and flow-based generative
models, using internal features, attention patterns, and hidden-state
perturbations to analyze how semantic content is introduced and refined
throughout generation~\cite{lin2025survey}.

In text-to-image models, generation unfolds over denoising steps and
repeatedly combines textual and visual information, making this perspective a
natural fit for analyzing how semantics emerge across the process.
Early work 
 adapts causal mediation analysis
to diffusion models and asks where visual attributes are stored across the text
encoder and U-Net~\cite{basu2023localizing}.
A subsequent line of work moves
from distributed attribute storage toward mechanistic localization,
seeking a small fraction of layers that exert especially strong control over
objects, style, or factual knowledge and can therefore support efficient model
editing~\cite{basu2024mechanistic}. Together, these studies suggest that text-to-image generators
are internally structured rather than homogeneous, and that different layers or
components can play distinct semantic roles.

More closely related to our setting, several recent works examine diffusion
transformers at the layer level. Stable Flow identifies
vital layers in flow-based DiTs and shows that selective feature
injection at those layers can support training-free image editing~\cite{avrahami2025stable}.
Recent MM-DiT analyses further report that only a subset of layers is strongly
responsible for text-image alignment or other downstream behaviors~\cite{lirevisiting}.
Taken together, these results point to a shared observation: layer importance
in modern diffusion transformers is highly non-uniform.

Our work builds on this layer-wise perspective, but focuses on stereotype
binding during generation. Rather than using layer localization primarily for
editing or knowledge modification, we use causal tracing to identify
\emph{semantic binding hubs} and then study how these hubs support cross-modal
bias formation and sparse deployment-time mitigation. In this way, our work
places layer-wise MM-DiT analysis in a fairness-oriented intervention setting.% \subsection{Mechanistic Analysis of MM-DiTs}

%% file: sections/threat_model.tex
\section{Problem Definition and Threat Model}

We study deployment-time bias mitigation for multimodal diffusion transformers
(MM-DiTs) under a practical deployment setting. In real-world systems,
text-to-image models are usually exposed through managed APIs or platform services, so
full retraining, large-scale data recollection, or maintaining many
fairness-specialized model variants can introduce substantial computational and
operational overhead. We therefore consider a defender who can
intervene during inference, but seeks to avoid retraining the full
text-to-image model.

\subsection{Problem Definition}

Let $G_{\theta}$ denote a deployed MM-DiT with fixed parameters $\theta$. Given
a prompt $p$ and seed $z$, the model generates an image
\begin{equation}
    x = G_{\theta}(p, z).
\end{equation}
Following prior work on debiasing evaluation in text-to-image generation, we
focus on occupation-related prompts following recent works~\cite{fu2025fairimagen, park2025fair,shi2025dissecting}, which provide a representative
setting in which normal requests can systematically trigger demographic
stereotypes. Our goal is to construct an inference-time intervention
$\mathcal{I}$ such that
\begin{equation}
    \tilde{x} = G_{\theta}^{\mathcal{I}}(p, z)
\end{equation}
reduces harmful stereotype bias while preserving prompt semantics, image
quality, and utility on bias-irrelevant prompts.

\subsection{Threat Model}

\noindent
\textit{Adversary.}
We consider a practical bias-exposure threat model in which normal user
prompts trigger harmful demographic stereotypes learned by the model.
The risk therefore arises from benign, naturally occurring inputs. 
Malicious adversaries can exploit the risk to diminish the model's reputation or cause harm to affected groups.

\noindent
\textit{Threat surfaces.}
We consider two threat surfaces. The first is the \emph{text side}: the text
encoder may encode biased associations, so that the resulting condition
embeddings place certain occupations closer to particular demographic
attributes. 
The second is the \emph{visual side}: the diffusion
backbone itself may favor majority-group visual realizations if it is
trained on biased data under maximum-likelihood-style objectives~\cite{yu2025bimodal}.

\noindent
\textit{Generality and practicality.}
The stereotype bias risk is not specific to a particular MM-DiT implementation, but rather arises from the underlying multimodal transformer architecture and training data.
Previous work has observed general bias risks across different MM-DiTs~\cite{lyu2025existing,vice2025quantifying,dehdashtian2025oasis}, and we also verify this in our experiments.
Since any user of a text-to-image service can trigger the risk through normal prompts, it is important to understand and mitigate it in a practical deployment setting.

\noindent
\textit{Defender.}
The defender is the model service provider or platform operator that deploys
the MM-DiT through a public-facing interface. The defender can access
inference-time hidden states and intermediate features of the deployed model,
and may apply lightweight interventions at selected layers or timesteps. 
However, due to computational overhead considerations, they may refrain from model redevelopment or repeated fairness-specific adaptation.
\input{figures/causal_tracing_overview}

%% file: figures/causal_tracing_overview.tex
\begin{figure}[t]
  \centering
  \includegraphics[width=\linewidth]{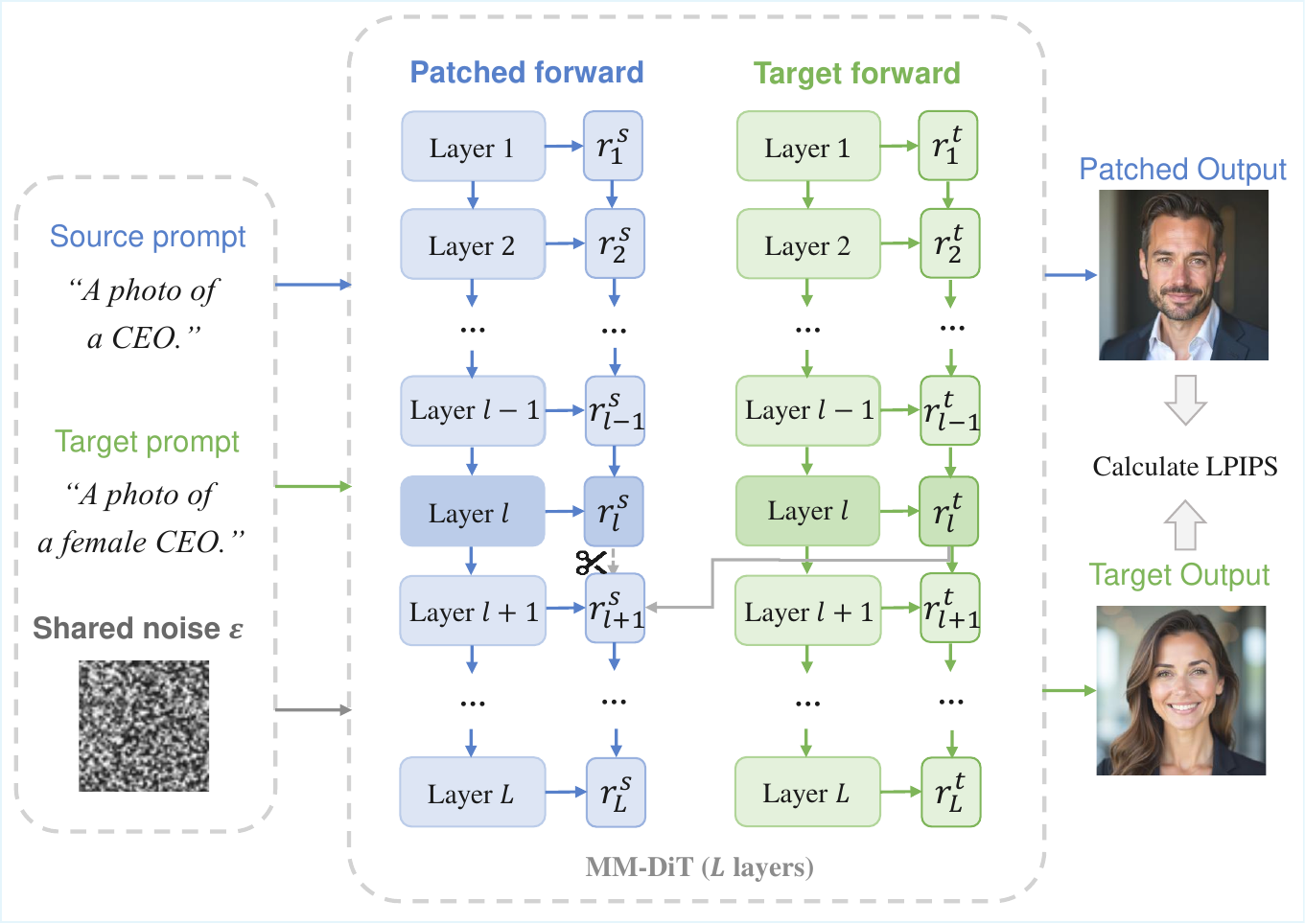}
  \caption{Residual causal tracing for locating hub layers. For each
  source--target prompt pair, we run the model with the same seed, replace the
  source residual output at a single layer with the corresponding target
  residual output, and compute the resulting causal score from
  $1-\mathrm{LPIPS}$ to the target generation.}
  \label{fig:causal_tracing_overview}
\end{figure}

%% file: sections/mechanistic_analysis.tex
\section{Root Cause Analysis of Bias Generation} \label{sec:analysis}

In this section, we identify how stereotype biases form and propagate during the generation process.
Our objective is to localize where demographic attributes acquire
causal influence during generation and to characterize how this influence is
propagated through the model. 

\subsection{Residual Causal Tracing for Locating Hub Layers}

To localize the internal layers that causally mediate stereotype formation, we
perform residual causal tracing based on layer-wise patching~\cite{wang2022interpretability}. 
The procedure is illustrated in Figure~\ref{fig:causal_tracing_overview}. For each
source--target prompt pair, the two prompts preserve the same occupational
concept while differing only in the demographic attribute of interest. We run
the model with the same random seed for both prompts and patch the source run
using the target run's residual output at a single layer within an early
timestep window. Importantly, this patching is applied only to the visual
tokens. The patched run is then compared with the target generation.
\textit{If patching a single layer makes the
source generation perceptually closer to the target generation, then that layer
must carry causal information for the swapped demographic attribute.}

Formally, let $x^{(s)} = G_{\theta}(p^{(s)}, z)$ and
$x^{(t)} = G_{\theta}(p^{(t)}, z)$ denote the source and target generations
under the same seed $z$. For a given layer $l$, we replace the source residual
output at that layer with the corresponding target residual output, yielding a
patched generation $x^{(s \leftarrow t)}_l$. We define the causal score of
layer $l$ as
\begin{equation}
    C(l) = 1 - \mathrm{LPIPS}\!\left(x^{(s \leftarrow t)}_l, x^{(t)}\right).
\end{equation}

We use $1-\mathrm{LPIPS}$ as the metric~\cite{zhang2018unreasonable}, where LPIPS measures perceptual image
distance, so that larger values indicate stronger target-side causal
influence.

Our tracing procedure relies on two modest but important assumptions. First,
for MM-DiTs with strong prompt fidelity, using the same seed and changing only
the attribute token typically keeps non-attribute semantics relatively
stable~\cite{wen2024detecting}. Second, demographic attributes are high-level semantic factors
rather than fine-grained texture details, so their key binding points tend to
emerge within an early portion of the generation trajectory~\cite{yi2024towards}. We therefore apply
residual patching only over the early time window $t \in [1.0, 0.7]$.

For each model, we instantiate attribute-swapped prompt pairs for both gender
and race while keeping the occupation fixed, compute layer-wise causal scores
for each pair, and average the resulting profiles across attribute settings.
Figure~\ref{fig:causal_hubs} summarizes the resulting layer-wise causal scores
for FLUX.1-dev and Stable Diffusion~3. In both models, the resulting profiles
are sparse and non-uniform rather than smoothly distributed across depth.

In FLUX.1-dev, the layer profile is sharply peaked, and we therefore identified the
dominant local maxima at layers 2, 18, 26, and 54. In Stable Diffusion~3, the
peaks are less isolated, so we select a small set of top-scoring layers,
namely 0, 1, 3, 19, 21, and 22. We refer to these consistently high-scoring layers as \emph{semantic binding hubs}.

\input{figures/causal_hubs}

\subsection{Functional Roles of Semantic Binding Hubs}

Having identified the hub layers by causal tracing, we next study whether these layers serve different semantic roles during generation. To achieve this, our analysis is primarily based on the internal attention mechanisms of the model. By examining the joint attention weights, we can directly observe how text and image tokens interact inside the generative backbone. We therefore analyze hub-layer behavior from two complementary views: image-to-text (I2T) and image-to-image (I2I). These views provide a unified picture of how stereotypical attributes are written, propagated, and consolidated across depth.

To systematically quantify these behaviors, we design specific metrics for both I2I and I2T interactions. For I2I analysis, we summarize each layer's spatial interaction pattern. Let $M_{\mathrm{I2I}}^{(l)} \in \mathbb{R}^{N}$ denote the aggregated attention map over image tokens at layer $l$, which quantifies the degree to which each image token is attended to by all other tokens across the entire image. Let $P^{(l)} = M_{\mathrm{I2I}}^{(l)} / \sum_i M_{\mathrm{I2I}}^{(l)}(i)$ be its normalized spatial distribution. We define the globality score as the normalized entropy:
\begin{equation}
G_{\mathrm{I2I}}^{(l)} = \frac{-\sum_i P^{(l)}(i)\log P^{(l)}(i)}{\log N},
\end{equation}
which measures how broadly the interaction is distributed over the image plane. Higher values indicate more global and spatially diffuse interactions. We also measure the overall I2I strength by the mean attention response:

\begin{equation}
S_{\mathrm{I2I}}^{(l)} = \frac{1}{N}\sum_i M_{\mathrm{I2I}}^{(l)}(i).
\end{equation}

For I2T analysis, we deliberately construct test prompts containing explicit attribute words and designate these words as the target tokens to quantify semantic write-in. Let $k^\star$ denote the target text token, $i$ an image token position, and $h$ an attention head. We first measure the attention probability from image tokens to the target word as:

\begin{equation}
A_{\mathrm{target}}^{(l)}(i,h) = \mathrm{softmax}\!\left(\frac{Q_{\mathrm{img}}^{(l)}K^{(l)\top}}{\sqrt{d}}\right)_{h,i,k^\star}.
\end{equation}
We then weight this attention by the norm of the target token's value vector,
\begin{equation}
E_h^{(l)}(i) = A_{\mathrm{target}}^{(l)}(i,h)\cdot \left\|V_{k^\star,h}^{(l)}\right\|_2,
\end{equation}
and aggregate across heads to obtain the target-token write-in score:
\begin{equation}
E^{(l)}(i) = \sqrt{\sum_h \left(E_h^{(l)}(i)\right)^2 }.
\end{equation}
This score measures how strongly image tokens attend to the target word and how much semantic content that word can inject through its value representation. A high score therefore indicates strong bias concept write-in from the text condition to the visual stream. 

With these metrics established, we now unpack the stage-wise functional roles of the semantic binding hubs.

\paragraph{Early hubs form coarse structure.}

The earliest hub layers primarily govern global structure formation. In FLUX.1-dev, this role is concentrated at layer 2, while in Stable Diffusion~3, it is most prominent at layers 0 and 1. As observed in the I2I heatmaps (Figure~\ref{fig:functional_hubs}), these early layers exhibit broadly distributed attention responses across the image plane, a phenomenon quantitatively reflected by their high $G_{\mathrm{I2I}}^{(l)}$ scores. As network depth increases within this initial stage, the attention responses progressively localize. This shift indicates a clear transition from drafting a global sketch to refining spatially focused structures, effectively establishing the foundational visual scaffold upon which later stereotypical attributes are written.

\paragraph{Middle hubs inject bias concepts.}

Middle hub layers play a distinct role focused on bias concept injection. As captured by the high target-token write-in scores $E^{(l)}(i)$, these layers are the most active sites for mapping textual concepts into visual features. Under our explicit attribute setup, layer 18 in FLUX.1-dev exhibits the strongest I2T write-in pattern toward the target attribute token; in Stable Diffusion~3, the corresponding behavior is concentrated at layer 3. At these specific layers, image tokens attend most strongly to the target word, receiving a highly concentrated semantic write-in signal. The causal tracing result is therefore perfectly complemented by this attention-based interpretation: these layers act as the primary bias concept injection hubs.

\paragraph{Late hubs integrate and broadcast global information.}

Late hub layers are associated with global bias-semantic consolidation. In FLUX.1-dev, this role is centered at layer 54, after which the model mainly broadcasts the written information through self-attention. In Stable Diffusion~3, a similar effect becomes apparent from layer 19 onward. At these depths, image-to-image (I2I) attention becomes globally distributed again. This global broadcasting behavior is quantitatively captured by a return to high $G_{\mathrm{I2I}}^{(l)}$ alongside a marked increase in $S_{\mathrm{I2I}}^{(l)}$. Rather than building the initial sketch, the model now propagates semantically bound content across the entire image. We also observe that layer 26 and its neighboring layers in FLUX.1-dev exhibit relatively weak I2I and I2T attention activity. Instead of relying on attention-driven spatial mixing, we hypothesize that the bias semantics bound at this depth are primarily processed and refined by the MLPs in the subsequent layers. This feature-level information is carried forward through the residual stream until it reaches the late broadcasting layers, where it is finally propagated to the entire image.

\input{figures/functional_hubs}
\paragraph{Non-hub layers.}
The contrast with non-hub layers is equally illuminating. In these intervening layers, attention activity is either strictly localized or largely dormant, indicating that their operations are dominated by local feature refinement and MLP processing rather than cross-modal attribute binding. This stark behavioral contrast confirms that semantic binding in MM-DiTs is not a uniformly distributed process, but rather a sparse, stage-wise collaboration. Early hub layers establish the structural scaffold, middle layers inject specific bias-semantic attributes, and late layers globally harmonize this accumulated information. Ultimately, this distinct division of labor highlights that cross-modal concepts including stereotype biases are anchored at highly specific structural bottlenecks, naturally paving the way for targeted mitigation.

\subsection{Cross-Modal Bias Reinforcement}

The hub-layer analysis suggests that bias generation in MM-DiTs is inherently
cross-modal, where visual tokens can absorb bias-associated semantics from text tokens.
We further show that this cross-modal interaction can create a feedback loop during generation: semantic attributes are not simply injected once from text to
image, but can be progressively reinforced through repeated multimodal
interaction. We examine this effect on Stable Diffusion~3 using prompt triplets
instantiated from the templates \texttt{\{occupation\}} (neutral), \texttt{\{bias-attribute\}+\{occupation\}}, and
\texttt{\{anti-bias attribute\}+\{occupation\}}. For each triplet, we run three forward passes with
the same seed and track the sentence-level text features throughout
generation.

At each timestep, we compare the neutral text features against the
corresponding bias and anti-bias trajectories. 
Since neutral prompts do not contain explicit attribute words, it shall not be strongly aligned with either bias or anti-bias trajectories if the generation process is purely feedforward. 
However, Figure~\ref{fig:cross_modal_reinforcement}(a)
shows that the average cosine similarity of the neutral text features has a clear separation early in
generation: it rapidly becomes more aligned with the
bias-associated trajectory than with the anti-bias trajectory. This indicates
that even when the prompt itself is neutral, the internal text representation
is quickly pulled toward the bias side due to the bias emerged in visual tokens.

To make this trend more explicit, we further define a bias axis at each
timestep using the text features of the bias and anti-bias prompts, and project
the neutral text features onto this axis. Figure~\ref{fig:cross_modal_reinforcement}(b)
shows that the neutral trajectory moves toward the bias side very early in the
generation process and continues to stay there across depth. We interpret this
as evidence of a cross-modal reinforcement loop: because joint attention allows
all tokens to attend to one another, the emerging visual state can feed back
into the text stream and further amplify bias-associated semantics.

This result reveals that bias in
MM-DiTs is not only written at sparse semantic binding hubs; it can also be
self-reinforced during generation through repeated multimodal interaction. In
other words, stereotype bias arises from an unfavorable internal closed loop
rather than from a one-shot text-side injection alone.

\input{figures/cross_modal_bias_reinforcement}

\input{figures/fairflow_overview}

%% file: figures/causal_hubs.tex
\begin{figure}[t]
  \centering
  \begin{minipage}[t]{0.48\linewidth}
    \centering
    \includegraphics[width=\linewidth]{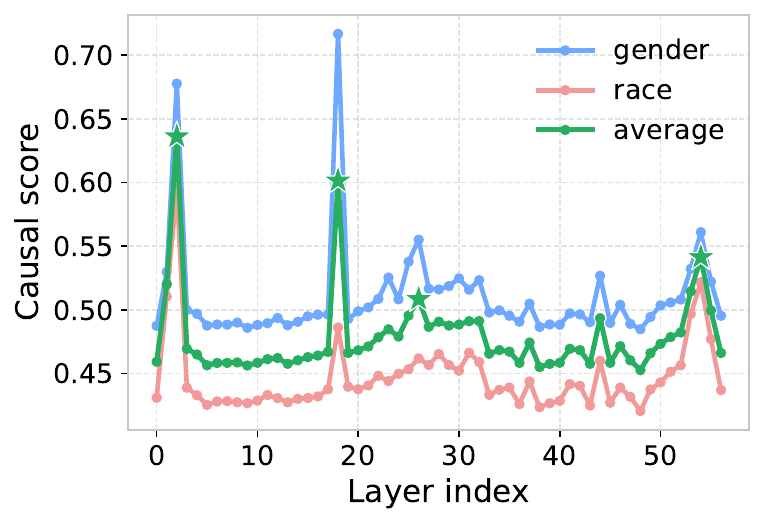}
  \end{minipage}\hfill
  \begin{minipage}[t]{0.48\linewidth}
    \centering
    \includegraphics[width=\linewidth]{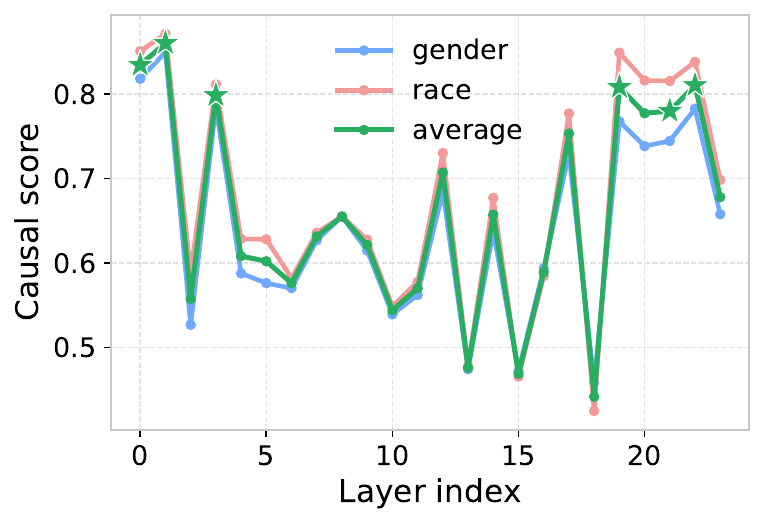}
  \end{minipage}
  \caption{Layer-wise causal scores from residual causal tracing on FLUX.1-dev
  (left) and Stable Diffusion~3 (right).}
  \label{fig:causal_hubs}
\end{figure}

%% file: figures/functional_hubs.tex
\begin{figure}[t]
  \centering
  \includegraphics[width=\linewidth]{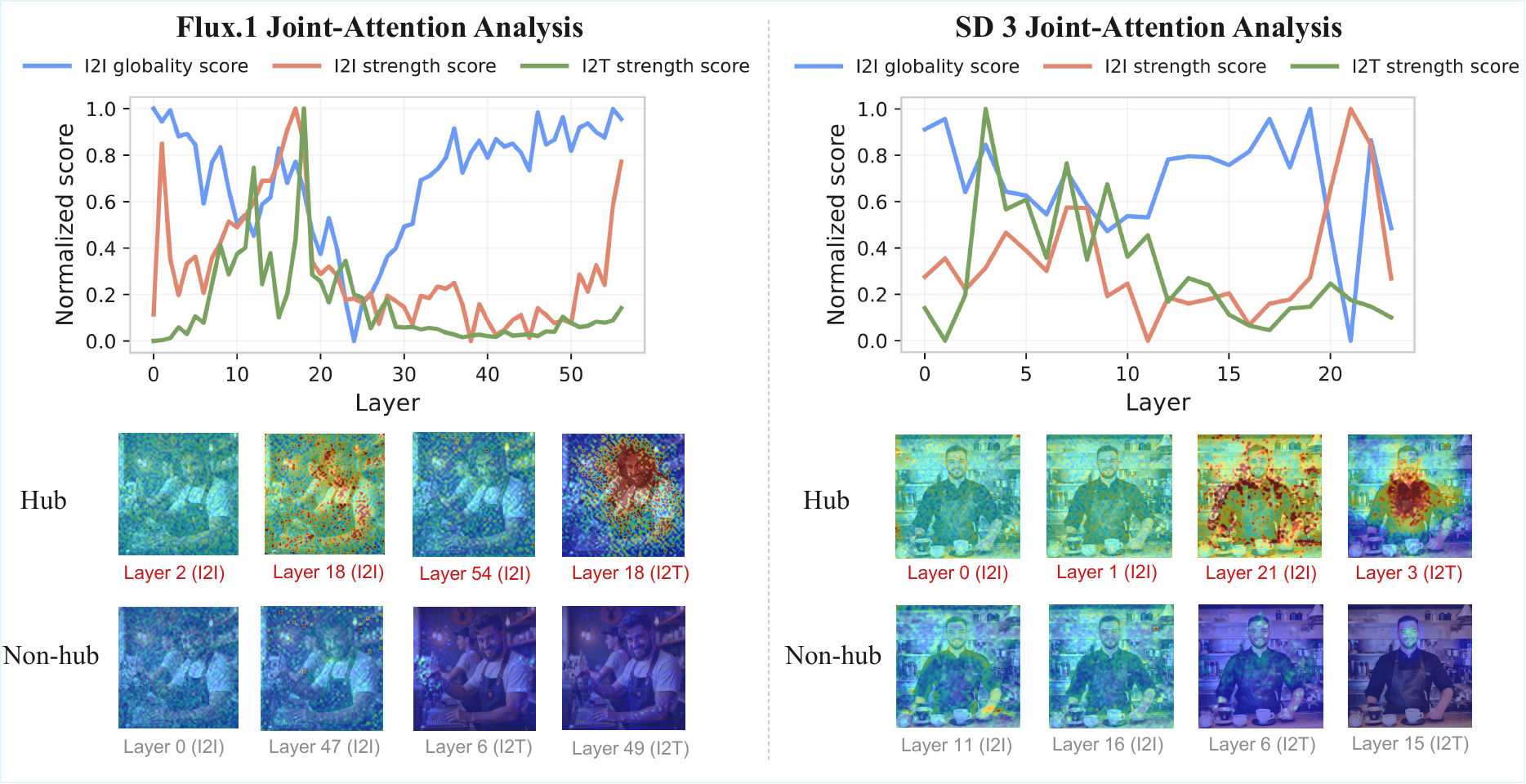}
  \caption{I2I and I2T attention analysis on FLUX.1-dev and Stable Diffusion~3, along with attention map visualizations of selected layers.}
  \label{fig:functional_hubs}
\end{figure}

%% file: figures/cross_modal_bias_reinforcement.tex
\begin{figure}[t]
  \centering
  \begin{minipage}[t]{0.48\linewidth}
    \centering
    \includegraphics[width=\linewidth]{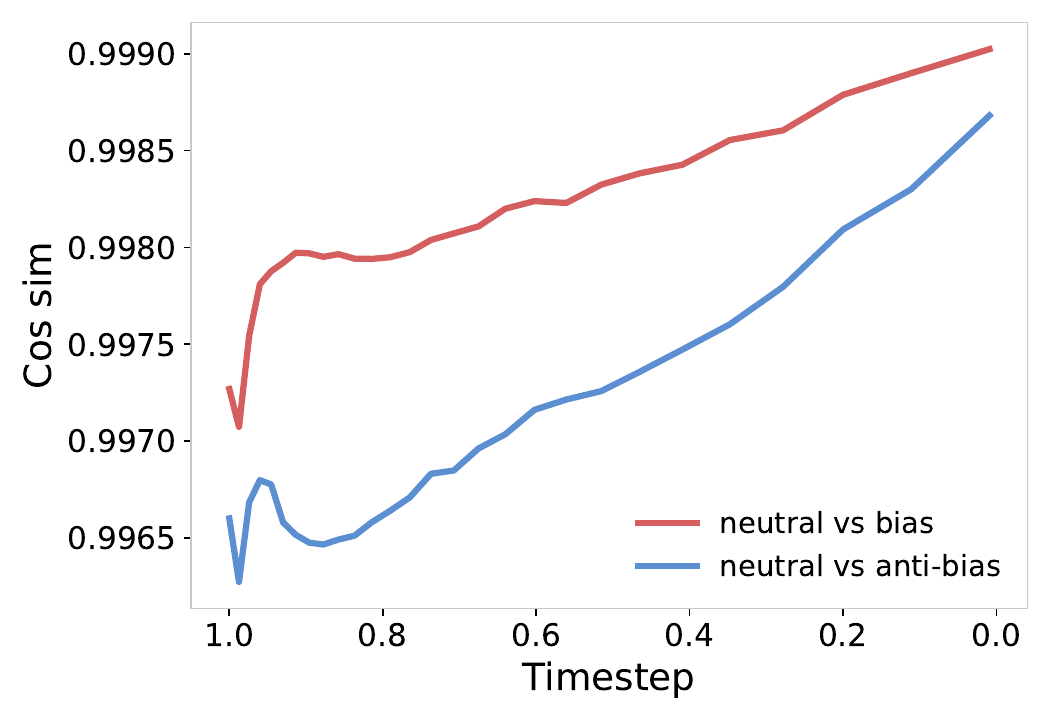}\\[-0.2em]
    \small (a) Cosine similarity
  \end{minipage}\hfill
  \begin{minipage}[t]{0.48\linewidth}
    \centering
    \includegraphics[width=\linewidth]{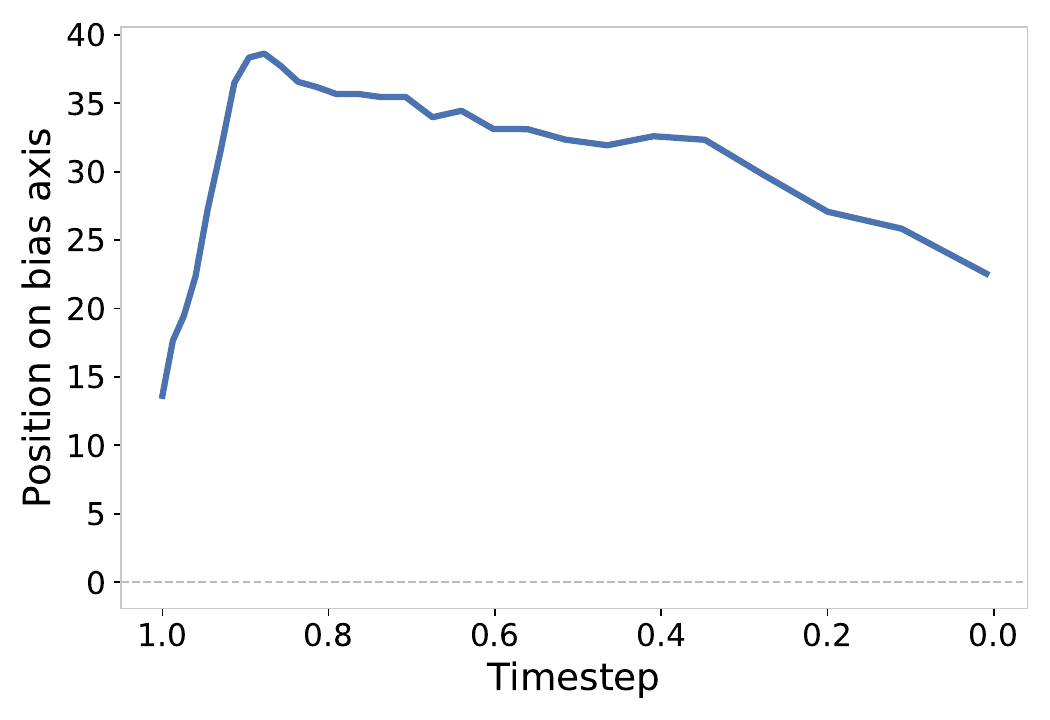}\\[-0.2em]
    \small (b) Bias-axis projection
  \end{minipage}
  \caption{Cross-modal bias reinforcement on Stable Diffusion~3. Using three
  forward passes with neutral, bias, and anti-bias prompts, (a) compares the
  average cosine similarity of the neutral text features to the bias and
  anti-bias text features across layers, and (b) shows the trajectory of the
  neutral text features projected onto the bias axis during generation.}
  \label{fig:cross_modal_reinforcement}
\end{figure}

%% file: figures/fairflow_overview.tex
\begin{figure*}[t]
  \centering
  \includegraphics[width=0.9\textwidth]{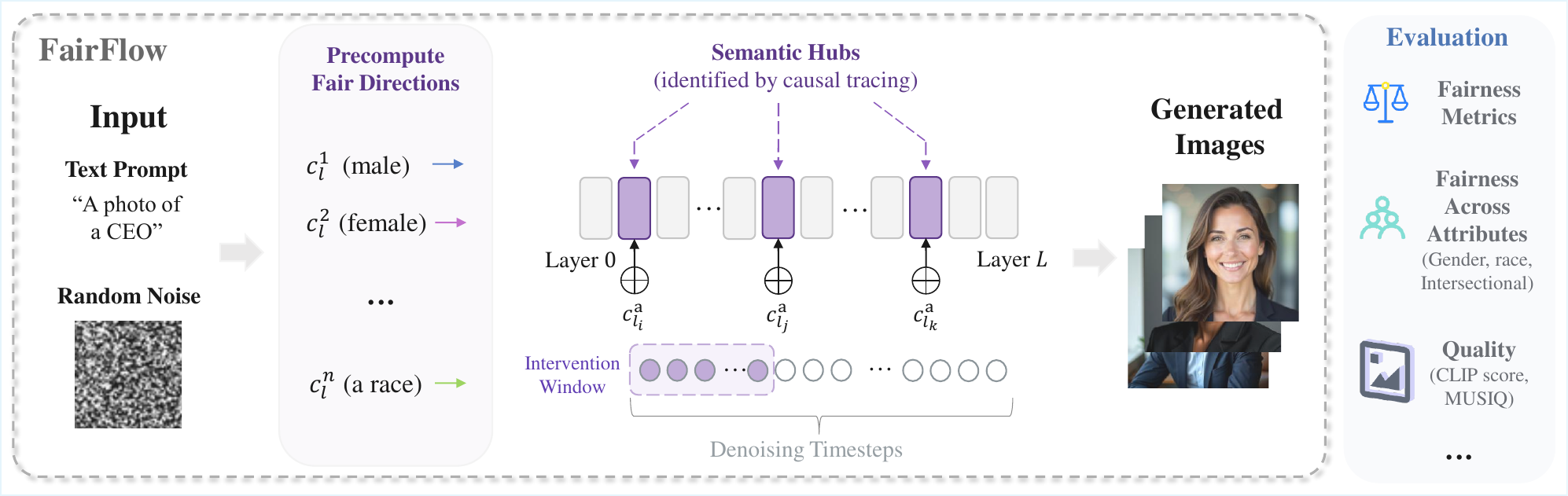}
  \caption{Overview of {\projname}. {\projname} precomputes attribute-specific fair
  directions, injects them only at the semantic binding hubs identified by
  causal tracing, restricts intervention to an early denoising window, and
  evaluates the resulting generations with fairness and quality metrics. The
  design is parameter-free at deployment time, efficient, and lightweight.}
  \label{fig:fairflow_overview}
\end{figure*}

%% file: sections/method.tex
\section{{\projname}: Sparse Steering at Semantic Binding Hubs}

Section~\ref{sec:analysis} shows that demographic attribute binding in MM-DiTs is concentrated
in a small number of semantic binding hubs which work together through multi-modal reinforcement to cause biased generations. 
In this section, we propose {\projname}, which turns the previous observation
into a deployment-time debiasing method. Rather than modifying the whole
model, we learn a small set of attribute-specific steering vectors and inject
them only at the identified hub layers. Figure~\ref{fig:fairflow_overview}
summarizes the full workflow of {\projname}, from precomputing fair directions to
sparse hub intervention and downstream evaluation. This design keeps the
intervention local, efficient, and easy to analyze.

\subsection{Steering Vector Optimization}
To learn a direction that can steer the model away from biased generations, we optimize steering vectors from paired prompts that share the same
occupational concept but differ in the demographic attribute of interest. For
each attribute, we construct source--target prompt pairs and learn one steering
vector per hub layer. The optimization is fully self-supervised: it uses only
the model's own internal features and denoising outputs, without requiring
additional human labels or external annotations.

Let $\mathcal{H}$ denote the set of hub layers identified in
Section~\ref{sec:analysis}. For each attribute $a$ and each layer $l \in \mathcal{H}$, {\projname}
learns a vector $c_l^{(a)} \in \mathbb{R}^{d}$ in the image hidden space. Given
a distribution of source--target prompt pairs $(p^{(s)}, p^{(t)}) \sim \mathcal{P}$ and initial noise seeds $z \sim \mathcal{Z}$, we sample intermediate noisy states $x_t$ and run the MM-DiT three times: a source forward pass, a target forward pass, and a steered source pass with additive steering vectors injected at the hub layers. Let $h_l^{(s)}$, $h_l^{(t)}$, and $\hat{h}_l$ denote the corresponding image features at hub layer $l$, and let $o^{(s)}$, $o^{(t)}$, and $\hat{o}$ denote the corresponding model outputs at the denoising step.

The optimization objective contains three terms. The first is a hub-matching
term. While conceptually this term aims to reproduce the source-to-target feature shift, mathematically it simplifies directly to minimizing the expected distance between the steered source features and the target features at the hub layers over the sampled batches:
\begin{equation}
    \mathcal{L}_{\mathrm{hub}}
    =
    \mathbb{E}_{p, z} \left[
    \frac{1}{|\mathcal{H}|}
    \sum_{l \in \mathcal{H}}
    \left\|
    \hat{h}_l - h_l^{(t)}
    \right\|_2^2
    \right].
\end{equation}
This term directly fits the causal layers identified in Section~\ref{sec:analysis} and ensures
that the learned vectors target the internal semantic binding process rather
than unrelated layers.

The second term matches the model output at the denoising step:
\begin{equation}
    \mathcal{L}_{\mathrm{out}}
    =
    \mathbb{E}_{p, z} \left[
    \left\|
    \hat{o} - o^{(t)}
    \right\|_2^2
    \right].
\end{equation}
In practice, this term is instantiated on the model's denoising prediction; for
rectified-flow-style models, it corresponds to aligning the predicted velocity
field. This auxiliary term complements hub matching by constraining the steered
trajectory at the model output level, which improves intervention quality and
stability.

The third term regularizes the vector magnitude:
\begin{equation}
    \mathcal{L}_{\mathrm{reg}}
    =
    \frac{1}{|\mathcal{H}|}
    \sum_{l \in \mathcal{H}}
    \left\|c_l^{(a)}\right\|_2^2.
\end{equation}
Without this constraint, directly fitting source--target differences tends to
absorb nuisance variation, such as background or lighting changes, which can
degrade generation quality. The regularizer keeps the learned vectors compact
and better aligned with attribute-specific changes.

The final training objective is
\begin{equation}
    \mathcal{L}
    =
    \lambda_{\mathrm{hub}} \mathcal{L}_{\mathrm{hub}}
    +
    \lambda_{\mathrm{out}} \mathcal{L}_{\mathrm{out}}
    +
    \lambda_{\mathrm{reg}} \mathcal{L}_{\mathrm{reg}}.
\end{equation}
These three terms play distinct roles. $\mathcal{L}_{\mathrm{hub}}$ makes the
vectors target the semantic binding hubs directly, $\mathcal{L}_{\mathrm{out}}$
encourages the steered trajectory to remain consistent with the target
generation, and $\mathcal{L}_{\mathrm{reg}}$ prevents overly large vectors that
would otherwise harm image quality. Empirically, using all three together gives
the best balance between debiasing strength and generation fidelity.

\subsection{Sparse Inference-Time Injection}

Once the vectors have been optimized, {\projname} applies them only at inference
time. Given an input prompt and an attribute direction $a$, we inject the
corresponding steering vectors only at the hub layers and only during the early
time window identified in Section~\ref{sec:analysis}, namely $t \in [1.0, 0.7]$. For a hub layer
$l$, the steered image features are first updated by
\begin{equation}
    \hat{h}_l = h_l + \alpha c_l^{(a)},
\end{equation}
where $\alpha$ is a scalar hyperparameter that controls the steering strength.
We then rescale the steered features back to the original feature norm,
\begin{equation}
    \tilde{h}_l
    =
    \frac{\|h_l\|_2}{\|\hat{h}_l\|_2}
    \hat{h}_l,
\end{equation}
which helps keep the hidden states on the model's native manifold and reduces
quality degradation during generation.

The concrete injection mechanism differs slightly across models. For Stable Diffusion~3, we inject the steering vectors into both the conditional and unconditional branches at the selected hub layers, which preserves the stability of classifier-free guidance (CFG). For FLUX.1-dev, since it defaults to CFG-free sampling, we directly inject the vectors into the image stream of the corresponding hub layers. In both cases, the intervention is restricted entirely to the visual tokens and leaves the text tokens unchanged.

\subsection{Design Principles}

% \noindent
\textit{Sparse over global intervention.}
{\projname} intervenes only at the hub layers identified in Section~\ref{sec:analysis}, rather
than modifying the full network.

% \noindent
\textit{Early-window intervention.}
The intervention is restricted to the early time window
$t \in [1.0, 0.7]$. Prior work on diffusion generation suggests that
high-level semantic attributes are largely established in the earlier part of
the trajectory, before later steps focus more on refinement and detail
adjustment~\cite{yi2024towards}.

% \noindent
\textit{Manifold stability.}
The training-time regularizer and inference-time norm preservation stabilize
the steered hidden states and help preserve generation quality.

Taken together, these principles make {\projname} a local rather than global
intervention. It introduces only minimal inference overhead, is easy to
attribute to a small set of interpretable hub layers, and is therefore better
suited to realistic deployment settings than methods that require model
parameter updates or broad interventions across many internal
layers~\cite{park2025fair,gandikota2024unified}. In deployment, the target group ratio can be adjusted by
choosing the corresponding attribute direction and its application frequency.

%% file: sections/implementation.tex
\section{Implementation and Experimental Setup} \label{sec:implementation} \label{sec:implementation}

\textbf{Models \& Implementation.}
We instantiate {\projname} on two representative MM-DiT models, FLUX.1-dev\footnote{\url{https://huggingface.co/black-forest-labs/FLUX.1-dev}}
and Stable Diffusion~3 Medium\footnote{\url{https://huggingface.co/stabilityai/stable-diffusion-3-medium-diffusers}}.
All experiments are conducted on Ubuntu 24.04.1 LTS with Python 3.12.8 and
PyTorch 2.6.0+cu118. We optimize steering vectors with Adam using a learning
rate of $5\times10^{-4}$ for 30 epochs on FLUX.1-dev and 100 epochs on Stable
Diffusion~3. The loss weights are set to
$\lambda_{\mathrm{hub}} = 1.0$,
$\lambda_{\mathrm{out}} = 0.2$, and
$\lambda_{\mathrm{reg}} = 0.01$. At inference time, the steering strength for
single-attribute intervention on FLUX.1-dev is set to $7.0$ for gender and
$5.0$ for race, while the corresponding strength on Stable Diffusion~3 is
$3.0$. The strength for intersectional attribute intervention is set to $2.0$
for FLUX.1-dev and 3.0 for Stable Diffusion~3.

\noindent\textbf{Baselines.}
We evaluate {\projname} against the vanilla (unmodified) models and four representative debiasing baselines, categorized by their intervention mechanisms.

\noindent\textit{\underline{Prompt-level methods}}.
For input-space interventions, we consider Fair Diffusion~\cite{friedrich2023fair}, evaluating its strategy of appending auxiliary instructions to shift the generation toward targeted attributes.

\noindent\textit{\underline{Embedding-level methods}}.
For methods operating in the fixed text embedding space, we evaluate FairImagen~\cite{fu2025fairimagen}, which projects text encodings toward a neutral representation, and compare against Weak Guidance~\cite{kim2025rethinking}, which explicitly injects target attribute embeddings into the condition vector.

\noindent\textit{\underline{Parameter-updating methods}}.
Finally, we compare against BiasNeuron~\cite{qi2025fine} to benchmark parameter-updating approaches. We evaluate this method because it similarly leverages mechanistic interpretability to localize bias, but requires costly fine-tuning rather than lightweight inference-time steering.

\noindent\textbf{Dataset.}
We follow the open-source evaluation setup of FairImagen~\cite{fu2025fairimagen,zhao2018gender}.
Specifically, we use 30 occupations to construct development prompts for hub
identification and steering-vector optimization, and other 100 occupations to
construct the test prompts for final evaluation. The basic prompt template is
\texttt{"a photo of a face of a \{occupation\}"}. In addition, we synthesize
50 complex-scene prompts with Gemini 3~\cite{team2023gemini} to evaluate whether {\projname}
remains effective in more diverse and realistic text-to-image settings.
For each prompt, we sample 6 different random seeds. To evaluate benign
utility on unrelated prompts, we further sample 200 prompts from the COCO 2017
validation set~\cite{lin2014microsoft} and generate one image for each prompt using randomly
sampled seeds.

\noindent\textbf{Metrics.}
We use two types of metrics. Fairness metrics evaluate whether the demographic
group proportions in generated images are balanced, while fidelity metrics
evaluate the semantic alignment and visual quality of the generated images.

\noindent\textit{\underline{Fairness metric}}.
We evaluate fairness by measuring the demographic group proportions in the
generated images. Following FairImagen~\cite{fu2025fairimagen}, we use a
lightweight pretrained facial attribute classifier from DeepFace~\cite{taigman2014deepface} to detect and
count members of each group in the generated images. Let $p$ denote the
vector of group proportions and $k$ the number of groups. The fairness score $F$
is defined as
\begin{equation}
    F = 1 - \frac{\sum_i \left|p_i - \frac{1}{k}\right|}{2\left(1 - \frac{1}{k}\right)}.
\end{equation}
Higher scores indicate more balanced group proportions. A score of 1 means
that all groups are generated at the same rate, while a score of 0 means that
only one group is generated.

\noindent\textit{\underline{Fidelity metrics}}.
We evaluate fidelity from two aspects: semantic alignment with the prompt and
intrinsic image quality. For text--image alignment, we use CLIP score~\cite{radford2021learning}
computed with CLIP ViT-B/16\footnote{\url{https://huggingface.co/openai/clip-vit-base-patch16}}. For image quality, we use MUSIQ~\cite{ke2021musiq}, a
transformer-based image quality assessment model that produces a perceptual
quality score for each generated image.

\noindent\textit{\underline{Tradeoff score}}.
To summarize the tradeoff between debiasing effectiveness and fidelity
preservation, we additionally define a fairness--fidelity tradeoff score.
We first compute a fidelity score by normalizing CLIP and MUSIQ
against the vanilla model and taking their geometric mean:
\begin{equation}
    U = \sqrt{
    \frac{\mathrm{CLIP}}{\mathrm{CLIP}_{\mathrm{vanilla}}}
    \cdot
    \frac{\mathrm{MUSIQ}}{\mathrm{MUSIQ}_{\mathrm{vanilla}}}
    }.
\end{equation}
We then combine $U$ with $F$ using the harmonic mean:
\begin{equation}
    T = \frac{2 F U}{F + U}.
\end{equation}
Higher values indicate a better balance between stereotype mitigation and
fidelity preservation.

%% file: sections/evaluation.tex
\section{Evaluation}

In this section, we evaluate {\projname} across two representative MM-DiT
architectures, FLUX.1-dev and Stable Diffusion~3, from six perspectives: bias mitigation on bias-prone prompts, utility preservation on bias-irrelevant prompts, robustness under more complex prompts, the effect on cross-modal bias reinforcement, the impact of key design choices, and deployment overhead. We organize the evaluation around the
following research questions:

\input{tables/flux3_debias_performance.tex}
\noindent\textbf{RQ1 [Debiasing Performance]:} How effective is {\projname} at
mitigating stereotype bias on bias-prone prompts across different MM-DiT
architectures?

\noindent\textbf{RQ2 [Utility Preservation]:} To what extent does {\projname}
preserve semantic alignment and image quality on bias-irrelevant prompts?

\noindent\textbf{RQ3 [Robustness]:} Does {\projname} remain effective when prompts
become more complex and realistic?

\noindent\textbf{RQ4 [Reinforcement Disruption]:} Does {\projname} weaken the internal cross-modal bias reinforcement identified in Section~\ref{sec:analysis}?

\noindent\textbf{RQ5 [Ablation Study]:} How do different design choices in {\projname} affect performance?

\noindent\textbf{RQ6 [Overhead]:} What additional inference overhead does {\projname} introduce in practice?

\subsection{RQ1: Debiasing Performance}

We first evaluate whether {\projname} effectively mitigates stereotype bias on
bias-prone prompts. This is the primary objective of our method. We compare
{\projname} against the vanilla model and representative debiasing baselines on
both FLUX.1-dev and Stable Diffusion~3 using the fairness and utility metric described in
Section~\ref{sec:implementation}.

\noindent\textbf{[RQ1-1] Balanced group proportions.}
On FLUX.1-dev (Table~\ref{tab:flux_debias_performance}), {\projname} delivers the
strongest debiasing performance across all three demographic settings. It
achieves the best fairness on gender ($0.600$), race ($0.489$), and
gender$\times$race ($0.556$), and also obtains the best average fairness
($0.548$). This margin remains substantial over both the vanilla model and the
competing baselines, indicating that the mechanism-guided intervention is able
to steer FLUX toward much more balanced group proportions without relying on
global prompt rewriting or parameter updates.

On Stable Diffusion~3 (Table~\ref{tab:sd3_debias_performance}), the picture is
also strong. {\projname} achieves the best race fairness ($0.656$) and the best
average fairness ($0.582$), while remaining competitive on both gender and
intersectional debiasing. This is a favorable result for deployment, because
it shows that {\projname} does not merely improve a single protected-attribute
setting, but instead delivers broad fairness gains across demographic axes on a
second MM-DiT architecture. Taken together with the FLUX results, this
demonstrates that the method generalizes well across MM-DiT families rather
than relying on model-specific behavior.

\noindent\textbf{[RQ1-2] Quality and semantics of debiased images.}
The quantitative results further highlight the inherent fairness--fidelity
tradeoff. On FLUX.1-dev, {\projname} secures the highest tradeoff scores across
all three settings and also the best average tradeoff score ($0.703$).
Crucially, it improves demographic balance without incurring the severe
generative degradation typical of competing baselines, overcoming the
traditional compromise between fairness and visual quality.

On Stable Diffusion~3, this tradeoff is more pronounced. FairDiffusion attains
the strongest gender fairness, but does so with large drops in CLIP and MUSIQ.
BiasNeuron preserves image quality well, yet yields only limited fairness gains
over the vanilla model. {\projname} strikes a stronger balance: it achieves the
best tradeoff on race ($0.789$), remains competitive on gender ($0.744$ vs.\
$0.687$ for FairDiffusion), and retains a strong intersectional tradeoff
($0.644$). Consequently, {\projname} achieves the highest average tradeoff
score ($0.727$, outperforming FairDiffusion's $0.608$ and WeakGuidance's
$0.492$), confirming its capacity to deliver the strongest overall
debiasing--fidelity balance even on challenging architectures.

\noindent\textbf{Take-away.} {\projname} consistently mitigates stereotype bias across both MM-DiT architectures while delivering the strongest overall fairness--fidelity tradeoff among all evaluated methods.
\subsection{RQ2: Utility Preservation}

A practical deployment-time debiasing method must not degrade core generation capabilities on bias-irrelevant prompts. To evaluate general utility preservation, we benchmark {\projname} against the vanilla model and established baselines using the COCO-based prompt set described in Section~\ref{sec:implementation}.

\input{figures/coco_fidelity}

As Figure~\ref{fig:coco_fidelity} illustrates, {\projname} delivers the strongest fidelity on FLUX.1-dev in both the gender and race settings ($0.994$) and ties with FairImagen in the intersectional setting ($0.976$). This demonstrates exceptional deployment readiness: the method successfully avoids the severe utility degradation typical of aggressive baselines like FairDiffusion, while matching the bias-irrelevant prompt fidelity of the most conservative embedding-level approaches.

On Stable Diffusion~3, {\projname} again substantially outperforms FairDiffusion, securing the highest fidelity in the gender and race settings ($0.998$ and $0.990$, respectively). Although FairImagen retains a slight statistical edge in the more complex intersectional setting ($0.975$ vs.\ $0.943$), {\projname} proves exceptionally robust overall. Taken together, these results confirm that {\projname} strictly preserves the model's original generation behavior, effectively bypassing the severe fidelity penalties incurred by less selective debiasing methods.

\noindent\textbf{Take-away.} {\projname} successfully preserves general model utility on bias-irrelevant prompts, establishing state-of-the-art fidelity preservation on FLUX.1-dev and remaining highly competitive on Stable Diffusion~3.

\input{tables/sd3_debias_performance.tex}
\subsection{RQ3: Robustness}

We next examine whether {\projname} remains effective when prompts contain richer
scene context and compositional constraints. To this end, we evaluate the
method on a complex-scene prompt set and measure both bias
mitigation and utility preservation under this more realistic setting. We
summarize the main fairness--fidelity trend in Figure~\ref{fig:complex_prompt_scatter}.

Figure~\ref{fig:complex_prompt_scatter} shows that the overall tradeoff pattern
remains favorable to {\projname} under complex prompts. On FLUX.1-dev, {\projname}
occupies the rightmost part of the plot while staying close to the top
fidelity region, indicating that it preserves scene semantics and image
quality substantially better than aggressive debiasing baselines at comparable
or stronger fairness levels. In particular, FairDiffusion can achieve larger
fairness gains in some settings, but these gains come with a visible drop in
fidelity, moving its points downward relative to the rest of the methods.

A similar pattern appears on Stable Diffusion~3. Most baselines cluster near
high fidelity but offer only limited fairness improvements, whereas
FairDiffusion again shifts toward higher fairness at the cost of a pronounced
fidelity collapse. {\projname} moves the operating point upward and to the right
relative to the vanilla model and the lighter baselines, yielding a stronger
balance between demographic fairness and generation fidelity across the three
demographic settings. This trend is especially important in the complex-scene
setting, where prompts contain substantial non-attribute semantic content and a
practical debiasing method must avoid degrading normal compositional control.

Overall, the scatter plot indicates that {\projname} remains robust when prompt
complexity increases: it continues to improve demographic balance while
preserving much more of the original fidelity than more aggressive baselines.

\noindent\textbf{Take-away.} {\projname} demonstrates consistent resilience when evaluated on complex prompts. It effectively enhances fairness over the vanilla model while securing a more favorable fairness--fidelity tradeoff than existing baselines.

\input{figures/complex_prompt_scatter}
\subsection{RQ4: Reinforcement Disruption}

Our mechanistic analysis in Section~\ref{sec:analysis} suggests that stereotype bias in MM-DiTs is sustained
by an internal cross-modal reinforcement process rather than a one-shot
text-side injection. We therefore ask whether {\projname} weakens this loop in
practice. To answer this question, we revisit the cosine similarity and bias-axis analysis from
Section~\ref{sec:analysis} and compare the source forward pass of the neutral prompt against the
corresponding steered forward pass produced by {\projname}.

Figure~\ref{fig:reinforcement_disruption} shows a clear change in the internal
text trajectory after steering. In Figure~\ref{fig:reinforcement_disruption}(a),
the source forward pass remains more aligned with the bias trajectory than with
the anti-bias trajectory, mirroring the cross-modal reinforcement effect in
Section~\ref{sec:analysis}. After applying {\projname}, the steered forward pass moves in the
opposite direction: its text features become less similar to the bias side and
more similar to the anti-bias side. Figure~\ref{fig:reinforcement_disruption}(b)
shows the same pattern from the bias-axis view. Without intervention, the
neutral trajectory drifts toward the bias side early in generation; with
visual-only steering, the trajectory is shifted back toward the anti-bias
side.

Crucially, although {\projname} intervenes exclusively on the visual stream, it fundamentally alters the subsequent text-side trajectory within the MM-DiT backbone. Our method therefore does not merely shift the final surface-level output distribution. Instead, it effectively weakens the internal reinforcement loop through which neutral prompts are otherwise pulled toward bias-associated semantics during the generative process.

\noindent\textbf{Take-away.} {\projname} actively disrupts the internal cross-modal bias reinforcement loop, demonstrating that our intervention neutralizes the bias formation process at its root, rather than merely adjusting surface-level outputs.

\input{figures/reinforcement_disruption}
\subsection{RQ5: Ablation Study}

Finally, we study the impact of key design choices in {\projname}. We focus on
three factors: the way steering vectors are constructed, the selected
intervention layers, and the steering strength. These ablations help clarify
whether the gains of {\projname} truly come from the mechanism-guided design.
\input{tables/sd3_ablation}
Table~\ref{tab:sd3_ablation} reveals three consistent patterns regarding our design choices. 

First, the optimization process is indispensable. In DirectDiff, the steering vector is computed directly as the hidden-state difference between the target forward pass and the source forward pass. While this simple construction preserves high CLIP and MUSIQ scores, its fairness performance remains stagnant, closely mirroring the vanilla model across all three settings. This demonstrates that directly extracting steering vectors from source--target feature differences is insufficient to produce meaningful debiasing. Consequently, the explicit optimization objective in {\projname} is necessary to align the intervention with the targeted semantic shift.

Second, targeted layer selection is critical. Although intervening at random layers yields fairness improvements, these gains consistently fall short of {\projname} and offer noticeably weaker control. 
Furthermore, restricting the intervention exclusively to the bottom layers proves even less effective. 
These empirical results firmly corroborate our mechanistic hypothesis: stereotype-relevant control is not uniformly distributed across network depth, but rather distinctly concentrated at the semantic binding hubs identified in Section~\ref{sec:analysis}.

Third, the steering strength $\alpha$ directly controls the fairness--fidelity tradeoff. Our default configuration already provides a strong general-purpose balance. When the intervention strength is smaller, the method preserves image quality more faithfully and remains closer to vanilla generation. Increasing $\alpha$ can further improve fairness scores, but excessively large values introduce visible quality loss. This suggests that $\alpha$ can be adjusted according to deployment needs, with the default setting serving as a balanced choice in practice.

\noindent\textbf{Take-away.} The ablations validate our core design: optimization is required for effective steering, targeted hub selection is superior to arbitrary layer placement, and steering strength provides a practical way to navigate the fairness--fidelity tradeoff.

\subsection{RQ6: Overhead}

Finally, we measure the deployment overhead introduced by {\projname} during
inference. We compare only methods that intervene at inference time, since
training-time approaches incur a different cost profile. Since {\projname}
intervenes only at a sparse set of hub layers within a limited inference
window, its additional cost should remain small relative to full generation.

\input{tables/sd3_time_overhead}

Table~\ref{tab:sd3_time_overhead} reports the additional inference overhead on
Stable Diffusion~3 for these inference-time intervention methods. {\projname}
introduces only $0.056$ seconds of extra runtime, which is still the smallest
overhead among all compared methods. This cost is lower than WeakGuidance
($0.166$ s), FairImagen ($3.240$ s), and especially FairDiffusion ($6.002$ s).

\noindent\textbf{Take-away.} {\projname} introduces negligible additional
inference overhead in practice, making it substantially lighter than competing
debiasing baselines.

%% file: tables/flux3_debias_performance.tex
\begin{table*}[t]
    \centering
    \caption{Debiasing performance on bias-prone prompts for FLUX.1-dev. We
    report fairness, CLIP score, and MUSIQ for gender, race, and
    intersectional gender$\times$race settings, together with the
    corresponding fairness--fidelity tradeoff scores and their averages. Higher
    is better for all metrics.}
    \label{tab:flux_debias_performance}
    \footnotesize
    \setlength{\tabcolsep}{1.8pt}
    \renewcommand{\arraystretch}{0.93}
    \setlength{\heavyrulewidth}{1.2pt}
    \resizebox{0.96\textwidth}{!}{%
    \begin{tabular}{>{\raggedright\arraybackslash}m{2.1cm}cccccccccccccccc}
    \toprule
    \multicolumn{1}{c}{\multirow{2}{*}{Method}} & \multicolumn{4}{c}{Gender} & \multicolumn{4}{c}{Race} & \multicolumn{4}{c}{Gender$\times$Race} & \multicolumn{4}{c}{Average} \\
    \cmidrule(lr){2-5}\cmidrule(lr){6-9}\cmidrule(lr){10-13}\cmidrule(lr){14-17}
    & Fairness & CLIP & MUSIQ & \cellcolor{gray!15}Tradeoff & Fairness & CLIP & MUSIQ & \cellcolor{gray!15}Tradeoff & Fairness & CLIP & MUSIQ & \cellcolor{gray!15}Tradeoff & Fairness & CLIP & MUSIQ & \cellcolor{gray!15}Tradeoff \\
    \midrule[0.8pt]
    Vanilla & 0.084 & \textbf{0.712} & \underline{0.735} & \cellcolor{gray!15}0.155 & 0.164 & \textbf{0.712} & \underline{0.735} & \cellcolor{gray!15}0.282 & 0.135 & \textbf{0.712} & 0.735 & \cellcolor{gray!15}0.238 & 0.128 & \textbf{0.712} & \underline{0.735} & \cellcolor{gray!15}0.226 \\
    \cmidrule(l){1-17}
    FairDiffusion & 0.244 & 0.685 & \textbf{0.740} & \cellcolor{gray!15}0.391 & \underline{0.464} & 0.683 & \textbf{0.742} & \cellcolor{gray!15}\underline{0.631} & \underline{0.323} & 0.673 & \textbf{0.742} & \cellcolor{gray!15}\underline{0.485} & \underline{0.344} & 0.680 & \textbf{0.741} & \cellcolor{gray!15}\underline{0.509} \\
    \cmidrule(l){1-17}
    WeakGuidance & \underline{0.486} & 0.671 & 0.727 & \cellcolor{gray!15}\underline{0.647} & 0.254 & 0.677 & \underline{0.735} & \cellcolor{gray!15}0.403 & 0.286 & 0.651 & 0.728 & \cellcolor{gray!15}0.440 & 0.342 & 0.666 & 0.730 & \cellcolor{gray!15}0.505 \\
    \cmidrule(l){1-17}
    FairImagen & 0.236 & \underline{0.706} & 0.728 & \cellcolor{gray!15}0.381 & 0.224 & \underline{0.705} & 0.727 & \cellcolor{gray!15}0.365 & 0.210 & 0.705 & 0.728 & \cellcolor{gray!15}0.347 & 0.223 & \underline{0.705} & 0.728 & \cellcolor{gray!15}0.364 \\
    \cmidrule(l){1-17}
    BiasNeuron & 0.276 & 0.670 & 0.726 & \cellcolor{gray!15}0.429 & 0.245 & 0.670 & 0.726 & \cellcolor{gray!15}0.391 & 0.222 & \underline{0.707} & \underline{0.738} & \cellcolor{gray!15}0.363 & 0.248 & 0.682 & 0.730 & \cellcolor{gray!15}0.395 \\
    \cmidrule(l){1-17}
    Ours & \textbf{0.600} & 0.704 & 0.725 & \cellcolor{gray!15}\textbf{0.746} & \textbf{0.489} & 0.703 & 0.728 & \cellcolor{gray!15}\textbf{0.654} & \textbf{0.556} & 0.687 & 0.710 & \cellcolor{gray!15}\textbf{0.706} & \textbf{0.548} & 0.698 & 0.721 & \cellcolor{gray!15}\textbf{0.703} \\
    \bottomrule
    \end{tabular}%
    }
    \end{table*}

%% file: figures/coco_fidelity.tex
\begin{figure}[t]
  \centering
  \begin{minipage}[t]{0.48\linewidth}
    \centering
    \includegraphics[width=\linewidth]{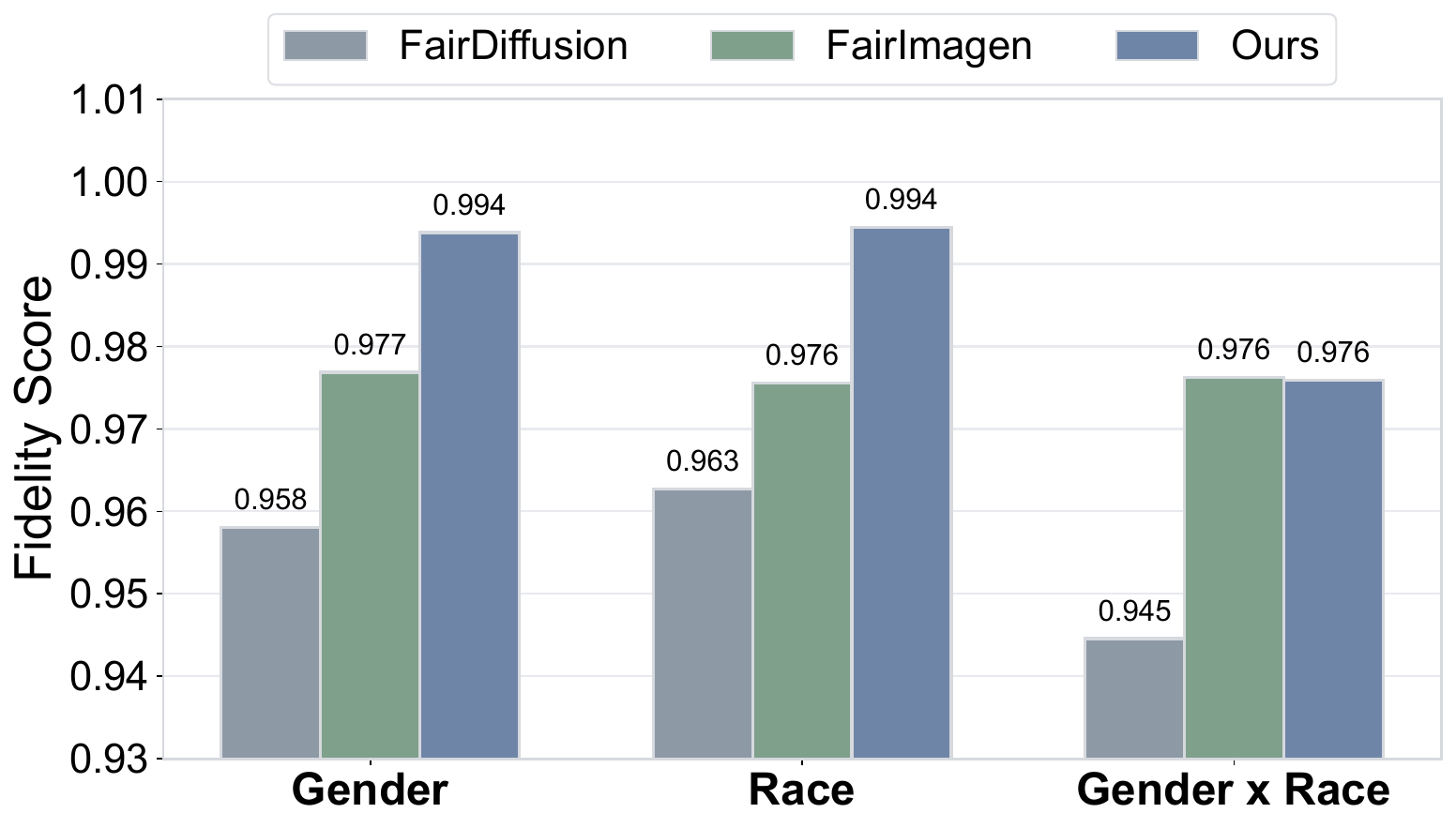}\\[-0.2em]
    \small (a) FLUX.1-dev
  \end{minipage}\hfill
  \begin{minipage}[t]{0.48\linewidth}
    \centering
    \includegraphics[width=\linewidth]{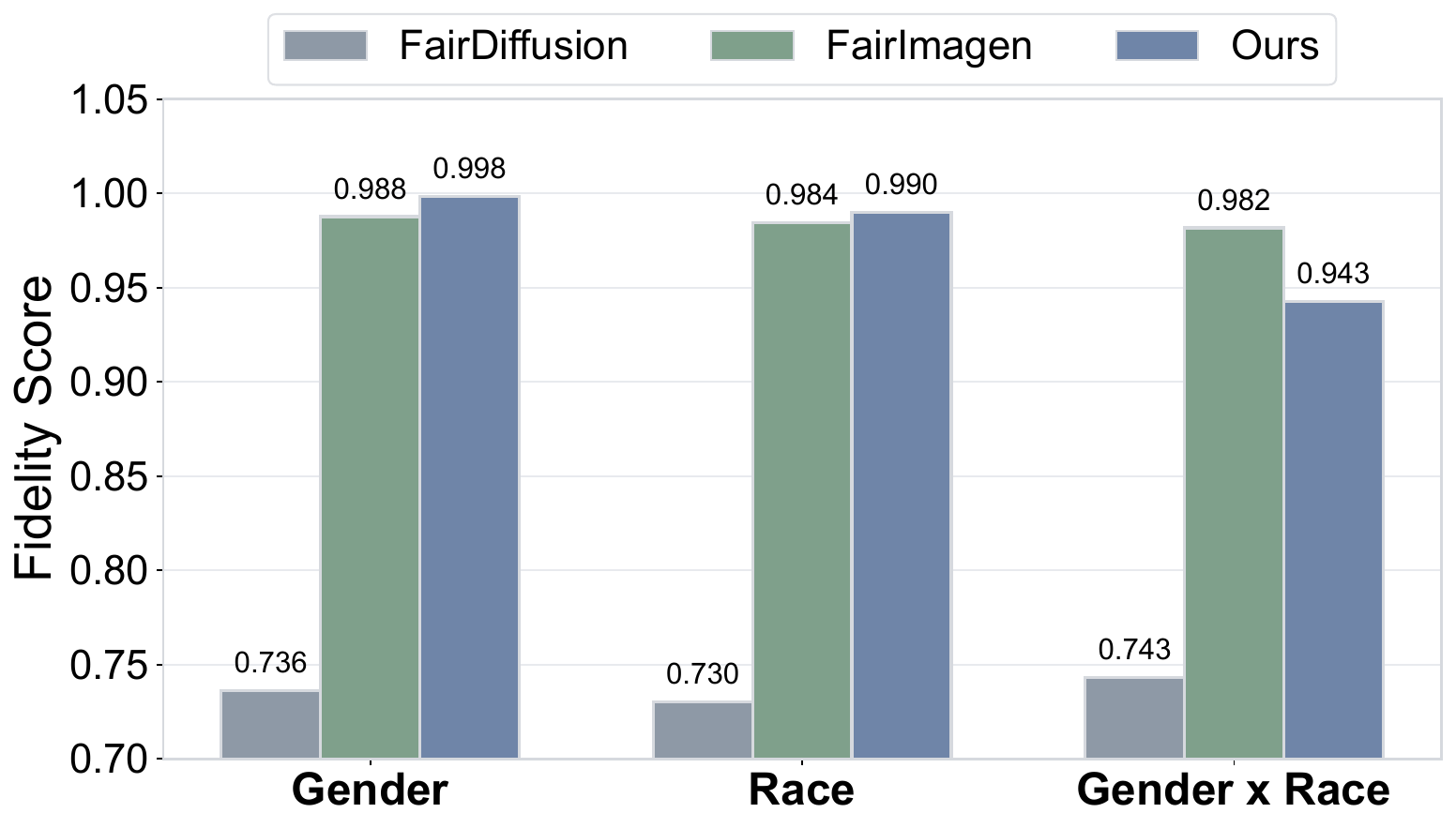}\\[-0.2em]
    \small (b) Stable Diffusion~3
  \end{minipage}
  \caption{Fidelity scores on bias-irrelevant COCO prompts.
  }
  \label{fig:coco_fidelity}
\end{figure}

%% file: tables/sd3_debias_performance.tex
\begin{table*}[t]
    \centering
    \caption{Debiasing performance on bias-prone prompts for Stable
    Diffusion~3. We report fairness, CLIP score, and MUSIQ for gender, race,
    and intersectional gender$\times$race settings, together with the
    corresponding fairness--fidelity tradeoff scores and their averages.
    Higher is better for all metrics.}
    \label{tab:sd3_debias_performance}
    \footnotesize
    \setlength{\tabcolsep}{1.8pt}
    \renewcommand{\arraystretch}{0.93}
    \setlength{\heavyrulewidth}{1.2pt}
    \resizebox{0.96\textwidth}{!}{%
    \begin{tabular}{>{\raggedright\arraybackslash}m{2.1cm}cccccccccccccccc}
    \toprule
    \multicolumn{1}{c}{\multirow{2}{*}{Method}} & \multicolumn{4}{c}{Gender} & \multicolumn{4}{c}{Race} & \multicolumn{4}{c}{Gender$\times$Race} & \multicolumn{4}{c}{Average} \\
    \cmidrule(lr){2-5}\cmidrule(lr){6-9}\cmidrule(lr){10-13}\cmidrule(lr){14-17}
    & Fairness & CLIP & MUSIQ & \cellcolor{gray!15}Tradeoff & Fairness & CLIP & MUSIQ & \cellcolor{gray!15}Tradeoff & Fairness & CLIP & MUSIQ & \cellcolor{gray!15}Tradeoff & Fairness & CLIP & MUSIQ & \cellcolor{gray!15}Tradeoff \\
    \midrule[0.8pt]
    Vanilla & 0.113 & \underline{0.707} & \textbf{0.749} & \cellcolor{gray!15}0.203 & 0.176 & \underline{0.707} & \textbf{0.749} & \cellcolor{gray!15}0.299 & 0.156 & \underline{0.707} & \textbf{0.749} & \cellcolor{gray!15}0.270 & 0.148 & \underline{0.707} & \textbf{0.749} & \cellcolor{gray!15}0.258 \\
    \cmidrule(l){1-17}
    FairDiffusion & \textbf{0.648} & 0.583 & 0.484 & \cellcolor{gray!15}\underline{0.687} & \underline{0.468} & 0.578 & 0.475 & \cellcolor{gray!15}\underline{0.567} & 0.452 & 0.560 & 0.504 & \cellcolor{gray!15}0.558 & \underline{0.523} & 0.574 & 0.488 & \cellcolor{gray!15}\underline{0.608} \\
    \cmidrule(l){1-17}
    WeakGuidance & 0.300 & 0.697 & \underline{0.748} & \cellcolor{gray!15}0.461 & 0.174 & 0.701 & \underline{0.747} & \cellcolor{gray!15}0.296 & \textbf{0.508} & 0.688 & \underline{0.739} & \cellcolor{gray!15}\textbf{0.669} & 0.327 & 0.695 & \underline{0.745} & \cellcolor{gray!15}0.492 \\
    \cmidrule(l){1-17}
    FairImagen & 0.173 & 0.682 & 0.719 & \cellcolor{gray!15}0.293 & 0.218 & 0.686 & 0.717 & \cellcolor{gray!15}0.356 & 0.174 & 0.684 & 0.717 & \cellcolor{gray!15}0.295 & 0.188 & 0.684 & 0.718 & \cellcolor{gray!15}0.315 \\
    \cmidrule(l){1-17}
    BiasNeuron & 0.128 & \textbf{0.710} & 0.738 & \cellcolor{gray!15}0.227 & 0.160 & \textbf{0.709} & 0.738 & \cellcolor{gray!15}0.276 & 0.124 & \textbf{0.710} & 0.737 & \cellcolor{gray!15}0.220 & 0.137 & \textbf{0.710} & 0.738 & \cellcolor{gray!15}0.241 \\
    \cmidrule(l){1-17}
    Ours & \underline{0.597} & 0.696 & 0.739 & \cellcolor{gray!15}\textbf{0.744} & \textbf{0.656} & 0.703 & 0.736 & \cellcolor{gray!15}\textbf{0.789} & \underline{0.494} & 0.679 & 0.670 & \cellcolor{gray!15}\underline{0.644} & \textbf{0.582} & 0.693 & 0.715 & \cellcolor{gray!15}\textbf{0.727} \\
    \bottomrule
    \end{tabular}%
    }
    \end{table*}

%% file: figures/complex_prompt_scatter.tex
\begin{figure}[t]
    \centering
    \includegraphics[width=0.95\linewidth]{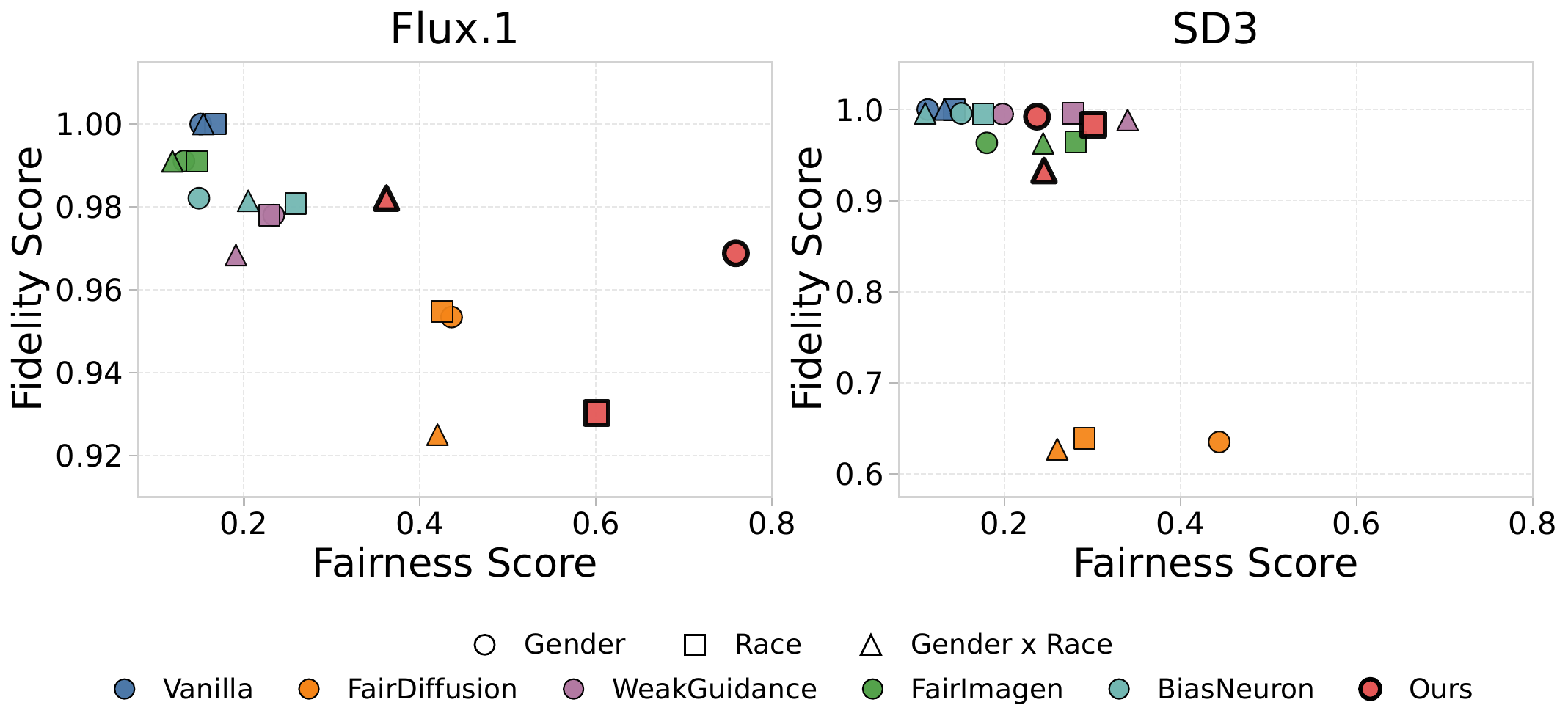}
    \caption{Fairness--fidelity tradeoff on complex-scene prompts. Each color
    corresponds to one method, with marker shape
    indicating gender, race, or gender$\times$race. 
    }
    \label{fig:complex_prompt_scatter}
\end{figure}

%% file: figures/reinforcement_disruption.tex
\begin{figure}[t]
  \centering
  \begin{minipage}[t]{0.48\linewidth}
    \centering
    \includegraphics[width=\linewidth]{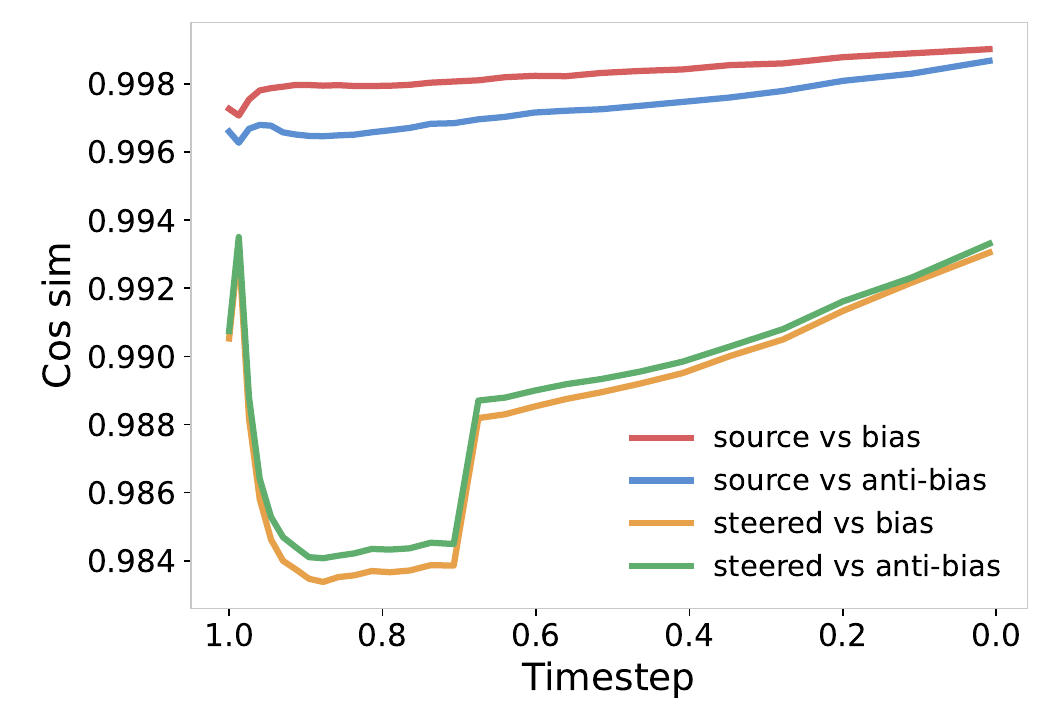}\\[-0.2em]
    \small (a) Cosine similarity
  \end{minipage}\hfill
  \begin{minipage}[t]{0.48\linewidth}
    \centering
    \includegraphics[width=\linewidth]{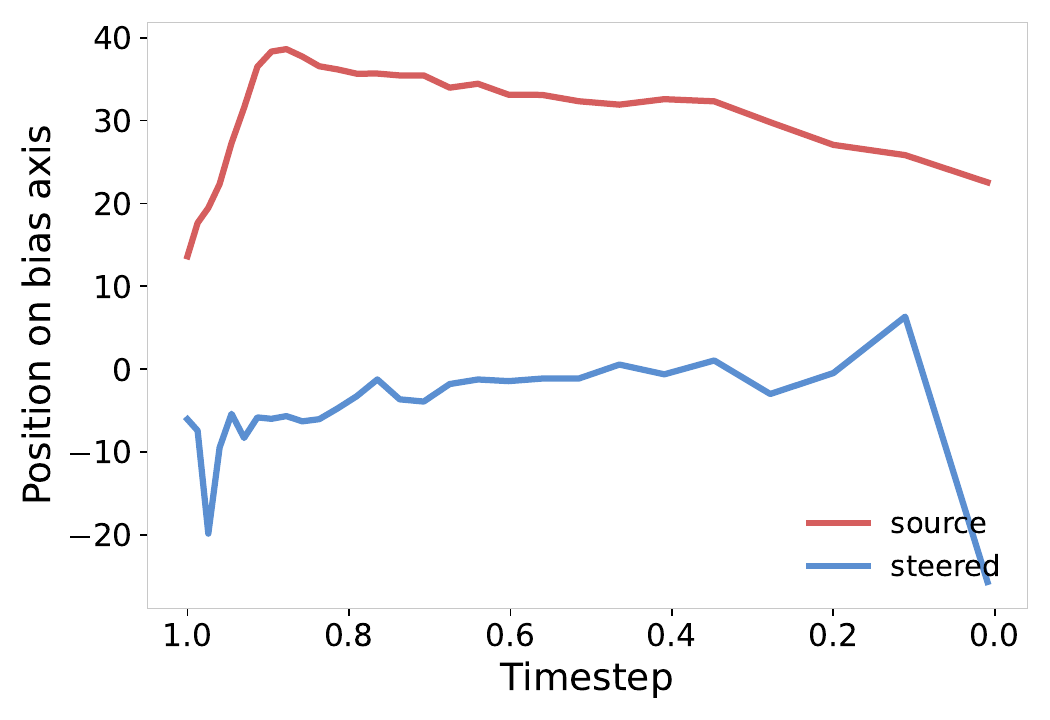}\\[-0.2em]
    \small (b) Bias-axis projection
  \end{minipage}
  \caption{Effect of {\projname} on cross-modal bias reinforcement. We compare the
  source and steered forward passes by measuring (a) their cosine similarity to
  the bias and anti-bias trajectories and (b) their positions on the bias axis
  during generation.
  }
  \label{fig:reinforcement_disruption}
\end{figure}

%% file: tables/sd3_ablation.tex
\begin{table}[t]
    \centering
    \caption{Ablation study on bias-prone prompts for Stable Diffusion~3. We
    compare vector construction, layer selection, and steering strength.}
    \label{tab:sd3_ablation}
    \setlength{\tabcolsep}{2pt}
    \setlength{\heavyrulewidth}{1.5pt}
    \resizebox{\columnwidth}{!}{%
    \begin{tabular}{>{\raggedright\arraybackslash}p{2.0cm}ccccccccc}
    \toprule
    \multicolumn{1}{c}{\multirow{2}{*}{Method}} & \multicolumn{3}{c}{Gender} & \multicolumn{3}{c}{Race} & \multicolumn{3}{c}{Gender$\times$Race} \\
    \cmidrule(lr){2-4}\cmidrule(lr){5-7}\cmidrule(lr){8-10}
    & Fairness & CLIP & MUSIQ & Fairness & CLIP & MUSIQ & Fairness & CLIP & MUSIQ \\
    \midrule[1pt]
    DirectDiff & 0.090 & 0.720 & 0.749 & 0.183 & 0.721 & 0.747 & 0.134 & 0.720 & 0.748 \\
    \midrule
    Random Layers & 0.430 & 0.713 & 0.749 & 0.506 & 0.717 & 0.742 & 0.398 & 0.702 & 0.743 \\
    \midrule
    Bottom Layers & 0.303 & 0.703 & 0.749 & 0.290 & 0.706 & 0.749 & 0.310 & 0.703 & 0.744 \\
    \midrule
    Low $\alpha$ & 0.467 & 0.697 & 0.721 & 0.396 & 0.702 & 0.736 & 0.414 & 0.687 & 0.701 \\
    \midrule
    High $\alpha$ & 0.621 & 0.685 & 0.656 & 0.598 & 0.694 & 0.723 & 0.336 & 0.618 & 0.551 \\
    \midrule
    {\projname} & 0.597 & 0.696 & 0.739 & 0.656 & 0.703 & 0.736 & 0.494 & 0.679 & 0.670 \\
    \bottomrule
    \end{tabular}%
    }
\end{table}

%% file: tables/sd3_time_overhead.tex
\begin{table}[t]
  \centering
  \caption{Additional inference overhead on Stable Diffusion~3 for
  inference-time intervention methods. Lower is better.}
  \label{tab:sd3_time_overhead}
  \footnotesize
  \setlength{\tabcolsep}{2.5pt}
  \setlength{\heavyrulewidth}{1.5pt}
  \resizebox{0.9\columnwidth}{!}{%
  \begin{tabular}{lcccc}
    \toprule
    Method & FairDiffusion & WeakGuidance & FairImagen & FairFlow \\
    \midrule[1pt]
    Overhead (s) & 6.002 & \underline{0.166} & 3.240 & \textbf{0.056} \\
    \bottomrule
  \end{tabular}%
  }
\end{table}

%% file: sections/discussion.tex
\section{Discussion}

{\projname} frames bias mitigation not as a resource-intensive data curation problem, but as a mechanism-guided internal defense. By operating directly at the identified semantic binding hub layers during inference, this localized, parameter-preserving intervention enables service providers to apply dynamic debiasing seamlessly. This approach entirely circumvents the need for complex prompt-rewriting heuristics or the computational burden of maintaining multiple fairness-specific model variants. Furthermore, the framework is inherently attribute-agnostic. While our current evaluation validates its efficacy on gender, race, and occupation, the underlying principle readily extends to a broader spectrum of social concepts, including complex intersectional identities and emerging cultural biases. By allowing flexible calibration of the intervention strength, {\projname} accommodates diverse user intents---ensuring robust fairness for underspecified queries while strictly preserving semantic adherence for fact-sensitive or explicitly descriptive inputs. Ultimately, this modular design can be seamlessly integrated with advancing automated fairness monitors, establishing a highly adaptable and computationally efficient paradigm for real-world content safety.

%% file: sections/conclusion.tex
\section{Conclusion}

In this work, we propose {\projname}, a deployment-time intervention framework designed to neutralize stereotype biases in multimodal diffusion transformers. Driven by our empirical discovery of stage-wise semantic binding, we demonstrate that cross-modal biases are deeply anchored at several specific layers. 
 By applying targeted, optimization-based steering exclusively at these crucial layers, our approach achieves state-of-the-art fairness--fidelity tradeoffs across diverse generative architectures. Our results further suggest that effective debiasing in MM-DiTs can be framed as weakening the cross-modal reinforcement loop that amplifies biased semantics during generation. {\projname} also offers a practical deployment path: it operates at inference time, requires no full-model retraining, and adds only minimal extra overhead. We expect that this work will inspire deeper investigations into the internal behaviors of generative models and serve as a foundation for safer and more equitable text-to-image systems.

%% file: sections/open_science.tex
\section{Open Science}

We will publicly release the code and artifacts needed to evaluate the core
contributions of this paper upon acceptance. The release will include the
{\projname} source code, evaluation scripts, prompt files, run scripts, and
documentation for reproducing the reported experiments.

We do not redistribute third-party foundation-model checkpoints or proprietary
dependencies. Instead, we provide the exact model identifiers, configuration
settings, and scripts needed to obtain the public artifacts from their original
sources. This choice reflects licensing, size, and distribution constraints.

We do not use human-subject data, private user data, or non-public demographic
annotations.

%% file: sections/ethics.tex
\section{Ethical Considerations}

This work studies stereotype bias in text-to-image generation and proposes a
deployment-time mitigation method. Its goal is harm reduction: to better
understand where bias forms inside MM-DiT systems and how it can be mitigated
without retraining the full model.

Several limitations are important. First, our evaluation uses simplified
gender- and race-based groupings as controlled experimental proxies; these
categories are not intended as a normative or exhaustive account of human
identity. Second, the evaluation relies in part on automated
attribute-recognition tools, whose outputs may themselves be imperfect or
biased, so the reported fairness scores should be interpreted as measurement
proxies rather than ground-truth social judgments. Third, FairFlow is designed
for stereotype-prone and underspecified prompts, not as a universal rewriting
mechanism for fact-sensitive or explicitly attribute-specified content.

Our released artifacts are intended to support reproducibility of a defensive
mechanism rather than any form of harmful profiling or identity inference. The
repository contains no private personal data, and the code is centered on
understanding and mitigating bias in generated outputs.

%% file: sections/generative_ai_usage.tex
\section{Generative AI Usage}

This paper was edited for grammar and light style polishing using GPT-5.4 and
Gemini 3 Pro. 

%% file: sections/supplementary_results.tex
\section{Additional Qualitative Results}

This appendix presents qualitative comparisons between the vanilla model and
{\projname} on two representative settings. We include both bias-irrelevant COCO prompts
and complex-scene prompts to illustrate whether {\projname} preserves normal
semantics while improving demographic balance.

\paragraph{Bias-irrelevant prompts.}
Figure~\ref{fig:appendix_coco_cases} compares vanilla generation and {\projname} on
COCO-based prompts for FLUX.1-dev and Stable Diffusion~3. These cases
provide a visual complement to RQ2 by showing that the intervention preserves
scene semantics and overall image quality on prompts that are not intended to
trigger stereotype bias.

\paragraph{Complex-scene prompts.}
Figure~\ref{fig:appendix_complex_cases} compares vanilla generation and
{\projname} on complex-scene prompts for FLUX.1-dev and Stable Diffusion~3. These
examples complement RQ3 by illustrating that {\projname} can still improve
demographic balance under richer compositional context without visibly
collapsing prompt fidelity.

\input{figures/appendix_coco_cases}
\input{figures/appendix_complex_cases}

%% file: figures/appendix_coco_cases.tex
\begin{figure*}[t]
    \centering
    \includegraphics[width=0.48\textwidth]{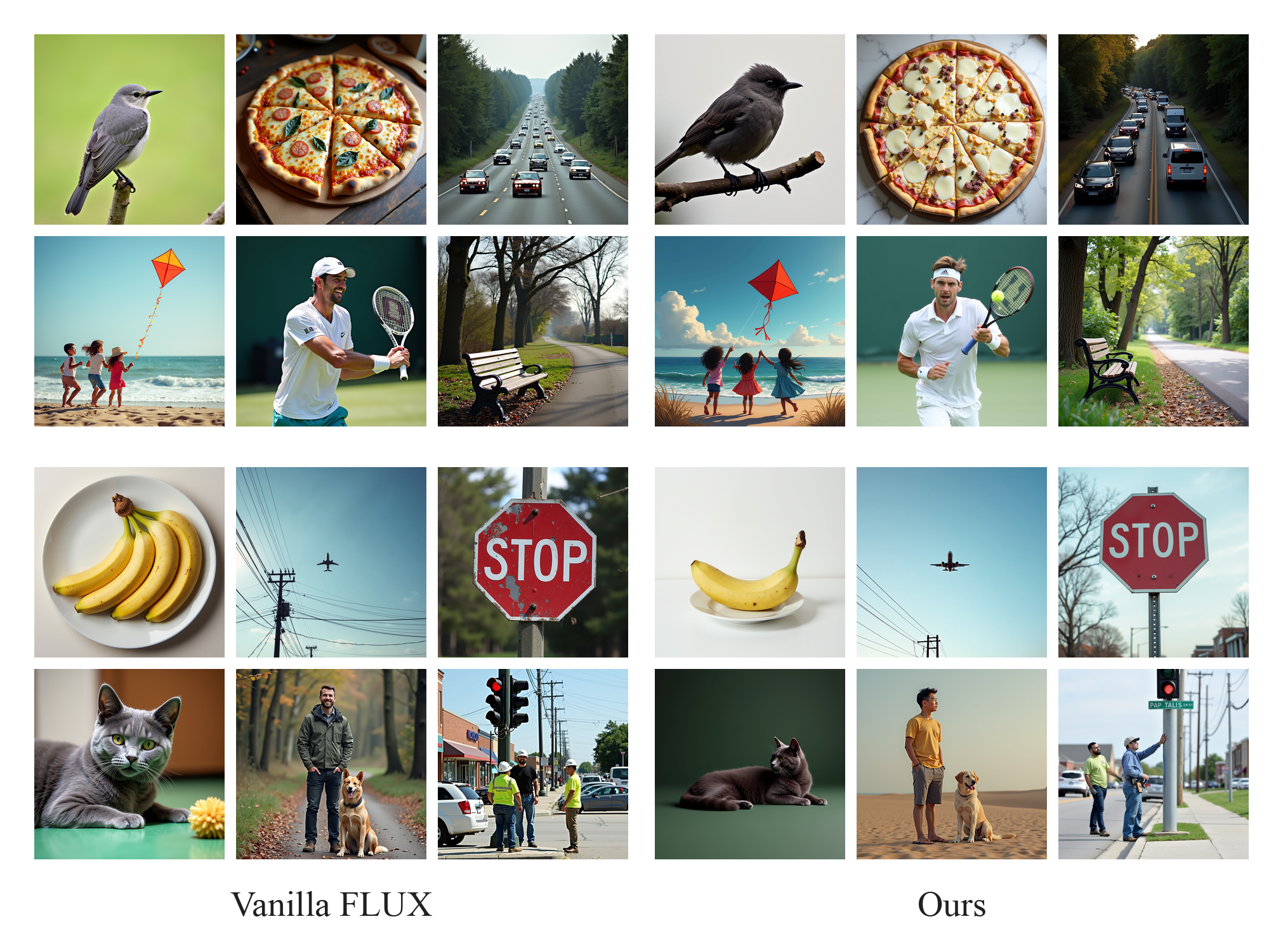}\hfill
    \includegraphics[width=0.48\textwidth]{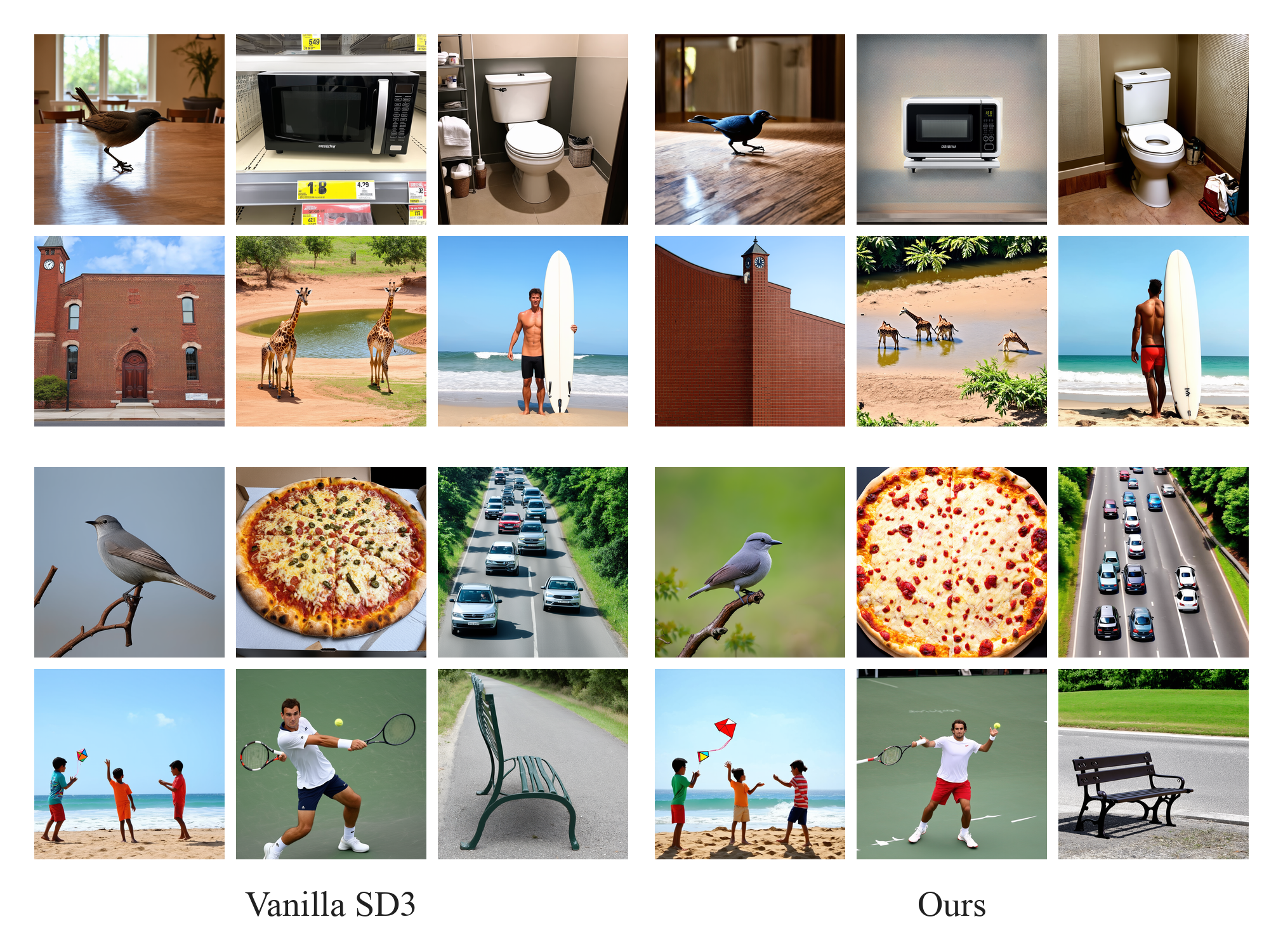}
    \caption{Qualitative comparison on bias-irrelevant COCO prompts. Left: FLUX.1-dev. Right: Stable Diffusion~3. In both models, {\projname} preserves the main scene semantics and visual quality while leaving unrelated generation behavior largely stable.}
    \label{fig:appendix_coco_cases}
\end{figure*}

%% file: figures/appendix_complex_cases.tex
\begin{figure*}[t]
    \centering
    \includegraphics[width=0.48\textwidth]{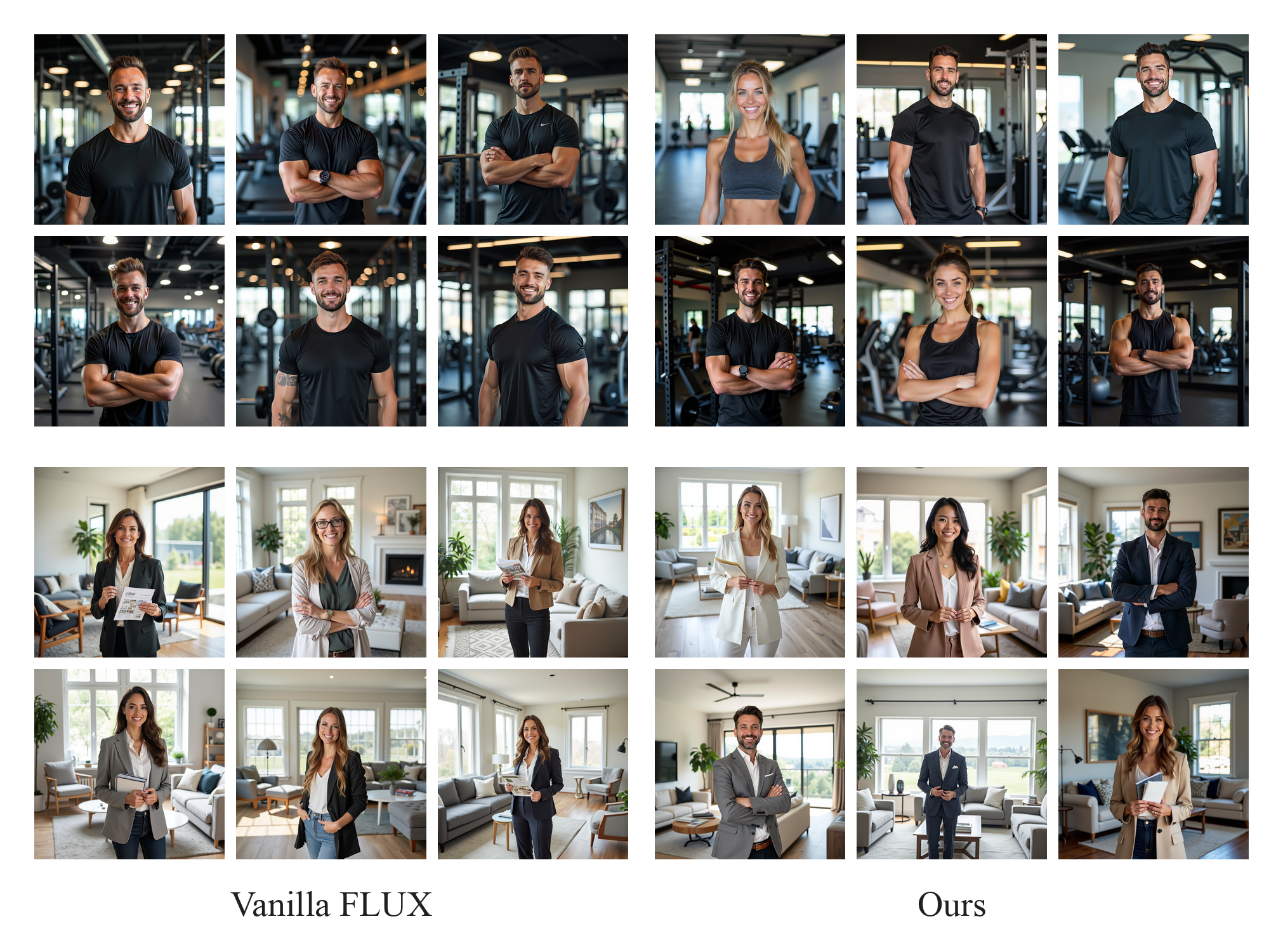}\hfill
    \includegraphics[width=0.48\textwidth]{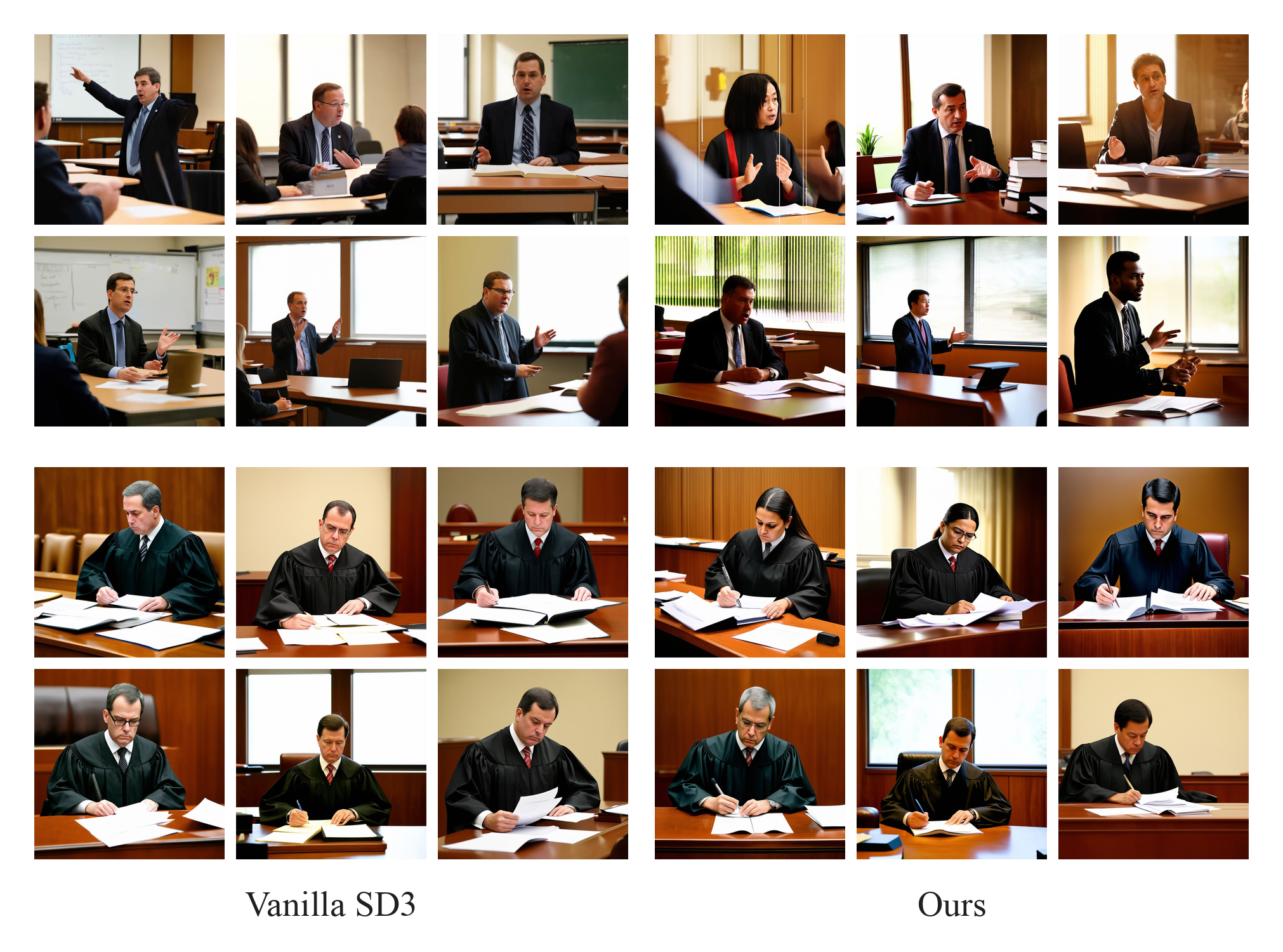}
    \caption{Qualitative comparison on complex-scene prompts. Left: FLUX.1-dev. Right: Stable Diffusion~3. {\projname} improves demographic balance while retaining the occupational, scene, and compositional content specified by the prompt.}
    \label{fig:appendix_complex_cases}
\end{figure*}